\documentclass[11pt,letterpaper]{article}
\usepackage[preprint]{jmlr2e}
\usepackage{amsmath,mathtools}
\usepackage{graphicx,booktabs}
\usepackage{enumitem}
\usepackage[ruled, linesnumbered]{algorithm2e}
\usepackage[capitalize,nameinlink]{cleveref}
\newcommand{\R}{\mathbb{R}}
\newcommand{\A}{\mathcal{A}}
\newcommand{\D}{\mathcal{D}}
\newcommand{\E}{\mathbb{E}}
\newcommand{\F}{\mathcal{F}}

\renewcommand{\H}{\mathcal{H}}
\newcommand{\X}{\mathcal{X}}
\newcommand{\Fk}{\mathcal{F}_{\kappa}}
\newcommand{\Ql}{\mathcal{Q}_{\lambda}}
\newcommand{\norm}[1]{\lVert #1\rVert}
\newcommand{\ip}[2]{\langle #1, #2\rangle}
\newcommand{\ALG}{\operatorname{\mathrm{ALG}}}
\newcommand{\OPT}{\operatorname{\mathrm{OPT}}}
\newcommand{\cost}{\mathrm{cost}}
\newcommand{\qs}{q^{*}}
\newcommand{\ds}{\delta^{*}}
\newcommand{\Cs}{C^{*}}
\DeclareMathOperator*{\argmin}{arg\,min}

\newcommand{\cb}{\mathbf{c}}
\newcommand{\x}{\mathbf{x}}
\newcommand{\y}{\mathbf{y}}
\newcommand{\z}{\mathbf{z}}
\newcommand{\vb}{\mathbf{v}}
\newcommand{\e}{\mathbf{e}}
\newcommand{\Uh}{\widehat{U}}

\newcommand{\ub}{\mathbf{u}}
\newcommand{\0}{\mathbf{0}}

\SetKwInput{KwInput}{Input}
\SetKw{Observe}{Observe}
\SetKw{Choose}{Choose}
\SetKw{Play}{Play}
\DontPrintSemicolon

\newcommand{\given}{\:|\:}
\newcommand{\Given}{\:\middle|\:}

\newtheorem{assumption}{Assumption}
\crefname{assumption}{Assumption}{Assumptions}
\Crefname{assumption}{Assumption}{Assumptions}

\usepackage{lastpage}

\ShortHeadings{Optimal Competitive Ratios for Chasing Convex Functions with Quadratic Growth}{He, Yang, Jiang, Zhang}
\firstpageno{1}

\title{Optimal Competitive Ratios for Chasing\\ Convex Functions with Quadratic Growth}
\author{\name {Linzhe He}\thanks{These authors contributed equally.}
	\email helz@lamda.nju.edu.cn \\
	\name {Sifan Yang}\footnotemark[1]
	\email yangsf@lamda.nju.edu.cn \\
	\addr State Key Laboratory of Novel Software Technology, Nanjing University \\
	\addr School of Artificial Intelligence, Nanjing University \\
	\AND
	\name {Wei Jiang}
	\email 12025220@njust.edu.cn \\
	\addr School of Computer Science and Engineering, Nanjing University of Science and Technology\\
	State Key Laboratory of Novel Software Technology, Nanjing University\\
	\AND
	\name {Lijun Zhang}
	\email zhanglj@lamda.nju.edu.cn \\
	\addr State Key Laboratory of Novel Software Technology, Nanjing University\\
	\addr School of Artificial Intelligence, Nanjing University\\
}
\date{}
\editor{}

\begin{document}

\maketitle

\begin{abstract}
Chasing convex function has attracted much attention in recent years, which can be formulated as a game  between a learner and an adversary. In each round, the adversary reveals a convex function, and the learner selects a decision and incurs a hitting cost  and a movement cost for changing its decision. The performance of the learner is measured by the \emph{competitive ratio}, which compares its cumulative cost with that of an optimal offline decision sequence. For Euclidean movement, existing upper and lower bounds for $\kappa$-well-conditioned functions exhibit a polynomial gap. For squared Euclidean movement, dimension-free competitive guarantees have been established under quadratic growth, but the optimal competitive ratio remains unclear.

In this paper, we establish optimal dimension-free competitive ratios for convex function chasing under two types of movement costs. For Euclidean movement, we propose Clipped Proximal Descent (CPD), which combines a balanced proximal step with a clipped correction toward the current minimizer. CPD achieves an $O(\kappa^{1/3})$ competitive ratio for smooth convex functions satisfying quadratic growth, improving the previous $O(\sqrt{\kappa})$ upper bound and matching the existing $\Omega(\kappa^{1/3})$ lower bound. For squared Euclidean movement, we establish the exact optimal dimension-free competitive ratio for convex functions satisfying quadratic growth. Our result is based on a new tight competitive analysis of Regularized Online Balanced Descent (R-OBD) and a matching lower bound against randomized algorithms.
\end{abstract}

\begin{keywords}
Convex function chasing, quadratic growth, competitive ratio, online optimization, optimal guarantee
\end{keywords}

\section{Introduction}\label{sec:intro}

In this paper, we study convex function chasing, which is a fundamental problem in online optimization with applications in dynamic resource provisioning and online control~\citep{AGG20}. Convex function chasing can be formulated as a game between a learner and an adversary. In each round $t$, the adversary reveals a convex function $f_t(\cdot)$, and the learner selects a decision $\mathbf{x}_t\in\mathcal X\subseteq\mathbb R^d$. The learner then incurs a cost of $f_t(\mathbf{x}_t)+m(\mathbf{x}_t,\mathbf{x}_{t-1})$, where $f_t(\mathbf{x}_t)$ is the hitting cost and $m(\mathbf{x}_t,\mathbf{x}_{t-1})$ is the movement cost of changing the decision from $\mathbf{x}_{t-1}$ to $\mathbf{x}_t$. To measure the performance of online algorithms, we adopt the competitive ratio, defined as
\begin{equation*}
\mathrm{CR}
=\frac{\sum_{t=1}^T\left(f_t(\mathbf{x}_t)+m(\mathbf{x}_t,\mathbf{x}_{t-1})\right)}
{\min_{\mathbf{u}_1,\ldots,\mathbf{u}_T\in\mathcal X}\sum_{t=1}^T\left(f_t(\mathbf{u}_t)+m(\mathbf{u}_t,\mathbf{u}_{t-1})\right)},
\end{equation*}
where $\mathbf{u}_0=\mathbf{x}_0$. The competitive ratio measures the worst-case ratio between the learner's cumulative cost and the minimum total cost achievable by an offline decision sequence.

For general convex functions with Euclidean movement, i.e., $m(\mathbf{x},\mathbf{y})=\|\mathbf{x}-\mathbf{y}\|$, \citet{FL93} establish an $\Omega(\sqrt{d})$ lower bound, while subsequent studies~\citep{AGG20,Sel20} achieve an $O(\min\{d,\sqrt{d\log T}\})$ competitive ratio. To obtain dimension-free guarantees, \citet{AGGT20} leverage additional structural properties of the hitting costs. They demonstrate that neither strong convexity nor smoothness alone suffices and consider $\kappa$-well-conditioned functions, which are $\alpha$-strongly convex and $\kappa\alpha$-smooth. For this class of functions, they propose constrained Online Balanced Descent, which achieves an $O(\sqrt{\kappa})$ competitive ratio, whereas any online algorithm has a worst-case ratio of at least $\Omega(\kappa^{1/3})$. While they establish a better bound, there still exists a polynomial gap between the upper and lower bounds in their dependence on $\kappa$.

On the other hand, for squared Euclidean movement, i.e., $m(\mathbf{x},\mathbf{y})=\frac12\|\mathbf{x}-\mathbf{y}\|^2$, strong convexity alone suffices to achieve dimension-free competitive ratios. \citet{GLSW19} develop Regularized Online Balanced Descent (R-OBD), achieving the optimal competitive ratio of $\frac12(1+\sqrt{1+4/\alpha})$ for $\alpha$-strongly convex functions. To extend these guarantees beyond strongly convex functions, \citet{ZJLY21} consider hitting costs satisfying quadratic growth, a standard relaxation of strong convexity~\citep{NNG19,DL18}. They provide a competitive ratio of $1+4/\lambda$ under $\lambda$-quadratic growth and $1+2/\sqrt{\lambda}$ with the additional assumption of convexity.

In this paper, we establish optimal dimension-free competitive guarantees for convex function chasing with quadratic growth under both movement costs. For Euclidean movement, we propose Clipped Proximal Descent (CPD), which combines a balanced proximal step with a clipped correction toward the current minimizer. When each hitting cost $f_t(\cdot)$ is $L_t$-smooth and satisfies $\lambda_t$-quadratic growth with $L_t\leq\kappa\lambda_t$, CPD achieves a better $O(\kappa^{1/3})$ competitive ratio. 
This improves the previous $O(\sqrt{\kappa})$ upper bound and matches the existing $\Omega(\kappa^{1/3})$ lower bound, establishing the optimal $\Theta(\kappa^{1/3})$ dependence on $\kappa$ under a weaker assumption than strong convexity. For squared Euclidean movement, we determine the exact optimal dimension-free competitive ratio for convex functions satisfying \(\lambda\)-quadratic growth. We establish a tight competitive analysis of R-OBD and a matching lower bound against randomized algorithms, and further quantify the effect of relaxing strong convexity on the optimal competitive ratio.

\subsection{Main results}\label{sec:results}

We establish dimension-free competitive guarantees under \Cref{ass:domain,ass:cost}, where dimension-free means that the guarantees hold for all dimensions $d$ and feasible sets $\X$. Both results rely on the quadratic growth condition~\citep{ZJLY21}. Specifically, a hitting cost $f_t(\cdot)$ with minimizer $\mathbf{v}_t$ satisfies $\lambda_t$-quadratic growth defined as
\begin{equation} \label{eq:qg}
    f_t(\x)-f_t(\vb_t)
    \ge
	\frac{\lambda_t}{2}\|\x-\vb_t\|^2
\end{equation}
for all $\mathbf{x}$ in the feasbile domain. For Euclidean movement, $\mathbf{v}_t$ is a global minimizer and the condition holds on $\R^d$, whereas for squared Euclidean movement, $\mathbf{v}_t$ minimizes $f_t(\cdot)$ over $\X$ and the condition holds on $\X$.

\paragraph{Euclidean movement.}
For Euclidean movement, let $\Fk$ denote the function class satisfying \Cref{ass:domain,ass:cost,ass:euclidean-qg,ass:euclidean-smooth}, i.e., each hitting cost $f_t(\cdot)$ is $L_t$-smooth and satisfies $\lambda_t$-quadratic growth on $\R^d$, with $L_t\leq\kappa\lambda_t$. The parameters may vary across rounds. We propose Clipped Proximal Descent (CPD), summarized in \Cref{alg:cpd}, which combines a balanced proximal step with a clipped correction toward the current minimizer.

\begin{theorem}\label{thm:distance}
For any $\kappa\geq 1$, CPD achieves a competitive ratio of $\frac{32}{7}(1+\kappa^{1/3})$ on $\Fk$.
\end{theorem}

\Cref{thm:distance} improves the $O(\sqrt{\kappa})$ upper bound of \citet{AGG20} and matches their $\Omega(\kappa^{1/3})$ lower bound~\citep{AGG20}. The dependence on $\kappa$ is therefore optimal.

\paragraph{Squared Euclidean movement.}
For $m(\mathbf{x},\mathbf{y})=\frac12\|\mathbf{x}-\mathbf{y}\|^2$, let $\mathcal{Q}_\lambda$ denote the class of instances where each hitting cost $f_t(\cdot)$ attains its minimum at $\mathbf{v}_t\in\mathcal{X}$ and satisfies $\lambda$-quadratic growth on $\mathcal{X}$. Unlike in the Euclidean movement setting, we do not require smoothness of the hitting costs here. We employ Regularized Online Balanced Descent (R-OBD)~\citep{GLSW19} and develop a novel tight competitive analysis. We further construct an oblivious distribution of hitting costs to establish a matching lower bound against randomized online algorithms.

Let $C^*(\lambda)$ denote the optimal competitive ratio on $\mathcal{Q}_\lambda$. To characterize it, we define
\begin{equation} \label{eq:squared:Lambda:phi}
    \Lambda(q,\delta)
    =\frac{(1+q\delta^2)^2}{q(1-\delta^2)(q+\delta)},
    \qquad
    \phi(q)=\min_{0\leq\delta<1}\Lambda(q,\delta),
\end{equation}
where $q>0$. The function $\phi(q)$ is continuous and strictly decreasing. Let $q^*(\lambda)$ denote the unique positive solution to $\phi(q)=\lambda$. We establish the following matching upper and lower bounds.

\begin{theorem}\label{thm:squared}
R-OBD achieves a competitive ratio of $1+q^*(\lambda)$ on $\mathcal{Q}_\lambda$. Moreover, any randomized online algorithm has a competitive ratio of at least $1+q^*(\lambda)$ on this class.
\end{theorem}

Combining the upper and lower bounds in \Cref{thm:squared}, we obtain the exact optimal competitive ratio $C^*(\lambda)=1+q^*(\lambda)$. We further derive an explicit upper bound and characterize its asymptotic behavior.

\begin{corollary}\label{cor:squared}
The optimal competitive ratio satisfies
\begin{equation*}
    C^*(\lambda)\leq
    1+\min\left\{\frac{1}{\sqrt{\lambda}},
    \frac{3\sqrt{3}}{2\lambda}\right\}.
\end{equation*}
Moreover,
\begin{equation*}
    \lim_{\lambda\to 0^+}\sqrt{\lambda}\,C^*(\lambda)=1,
    \qquad
    \lim_{\lambda\to\infty}\lambda\big(C^*(\lambda)-1\big)
    =\frac{3\sqrt{3}}{2}.
\end{equation*}
\end{corollary}

Corollary~\ref{cor:squared} improves the previous upper bounds of $1+2/\sqrt{\lambda}$ and $1+4/\lambda$ established by \citet{ZJLY21}. Moreover, compared with the optimal competitive ratio for strongly convex functions~\citep{GLSW19}, our result quantifies the effect of relaxing strong convexity to quadratic growth.


\subsection{Key techniques}\label{sec:overview}

\paragraph{Euclidean movement.}
The analysis of Online Balanced Descent (OBD) in previous work~\citep{AGG20} bounds the cost of each round through a decrease in the potential $\norm{\mathbf{x}_t-\mathbf{u}_t}$, where $\mathbf{u}_t$ is the comparator. A balanced step incurs hitting and movement costs proportional to its length, but the decrease in the potential depends on the direction of movement. In particular, when the step is nearly orthogonal to the direction toward the comparator, the potential may decrease much less than the movement cost, limiting the progress guaranteed by the balanced step alone. To overcome this challenge, we exploit the quadratic growth condition, which bounds the distance between the comparator and the current minimizer in terms of the comparator's excess hitting cost. Consequently, a comparator with small excess hitting cost lies close to the minimizer, and moving toward the minimizer can provide additional progress even when the balanced step is poorly aligned with the comparator. Based on this property, we propose CPD, which supplements the balanced proximal step with a clipped correction toward the current minimizer. Our analysis combines the progress of these two steps and employs quadratic growth and smoothness to control the additional movement and the error introduced by the correction, establishing an $O(\kappa^{1/3})$ competitive ratio.

\paragraph{Squared Euclidean movement.}
For squared Euclidean movement, we develop a novel competitive analysis of R-OBD~\citep{GLSW19} under quadratic growth. The main challenge is that quadratic growth only relates the comparator's excess hitting cost to its distance from the minimizer, without providing the curvature inequality between the online decision and the comparator available under strong convexity. Consequently, existing analyses relying on strong convexity cannot be directly applied. To overcome this difficulty, we apply the first-order optimality condition of R-OBD to both the comparator $\mathbf{u}_t$ and the current minimizer $\mathbf{v}_t$, and combine the resulting inequalities with a quadratic potential. To verify its optimality, we further construct a matching lower bound against randomized algorithms based on the equality conditions of the upper-bound analysis.

\paragraph{Organization.} \Cref{sec:related} reviews related work, and \Cref{sec:prelim} states the assumptions and analytical tools. \Cref{sec:distance} proves \Cref{thm:distance}; \Cref{sec:upper} provide the upper and lower bounds of squared Euclidean movement. \Cref{sec:conclusion} concludes the paper. Supporting proofs are presented in \Cref{app:supporting}.

\section{Related work}\label{sec:related}

\paragraph{Chasing convex bodies and functions.}
\citet{FL93} introduce body chasing, establish an $\Omega(\sqrt{d})$ lower bound,
and give a constant-competitive algorithm for $d=2$.
\citet{BGKPSS15,AS18} prove the optimal competitive ratio for function chasing for $d=1$.
For nested sets ($S_t\subseteq S_{t-1}$) case, \citet{BBEKU18} establish a ratio depending only on $d$, and \citet{ABCGL19} improve it to $O(d\log d)$.
For arbitrary sets case, \citet{BLLS19} obtain the first $2^{O(d)}$ ratio and reduce function chasing in $\R^d$ to body chasing in $\R^{d+1}$.
Subsequently, \citet{AGGT20} obtain an $O(\min\{d,\sqrt{d\log T}\})$ ratio for body chasing, and \citet{Sel20} establish the same guarantee for both problems.
These results make general chasing possible in every fixed dimension,
but their guarantees depend on dimension $d$.
This motivates the search for stronger guarantees when the cost functions have additional structure.

\paragraph{Dimension-free guarantees.}
For Euclidean distance movement,
\citet{CGW18} achieve a constant competitive ratio for locally polyhedral hitting costs.
For strongly convex and smooth hitting costs with condition number at most $\kappa$,
\citet{AGG20} obtain an $O(\sqrt{\kappa})$ ratio, and a lower bound $\Omega(\kappa^{1/3})$ for all online algorithms.
For squared Euclidean movement $m(\x,\y)=\frac12\|\x-\y\|_2^2$ and $\lambda$-strongly convex hitting costs,
\citet{GW19} establish a $3+O(1/\lambda)$ ratio.
Later, \citet{GLSW19} attain the optimal ratio of $(1+\sqrt{1+4/\lambda})/2$ for the $\lambda$-strongly convex hitting costs.

\paragraph{Quadratic growth condition.}
Quadratic growth is a weaker assumption than strong convexity and has been widely used to establish linear convergence in static convex optimization~\citep{NNG19,DL18}.
For squared Euclidean movement and quasi-convex hitting costs satisfying $\lambda$-quadratic growth,
\citet{GLSW19} prove that G-OBD is $O(\lambda^{-1/2})$-competitive as $\lambda$ approaches $0$.
\citet{LGW20} establish the ratio of $\max\{1+6/\lambda,4\}$ under $\lambda$-quadratic growth.
\citet{ZJLY21} improve this bound to $1+4/\lambda$,
and obtain a $1+2/\sqrt{\lambda}$ ratio for convex hitting costs.

\section{Preliminaries}\label{sec:prelim}

\paragraph{Notation.}
$\norm{\cdot}$ and $\ip{\cdot}{\cdot}$ denote the Euclidean norm and the standard inner product, respectively.
For a nonempty closed convex set $S\subseteq\R^d$, let $\Pi_S$ denote Euclidean projection onto $S$.
We introduce the following standard assumptions \citep{AGGT20}.

\begin{assumption}\label{ass:domain}
The decision set $\X\subseteq\R^d$ is nonempty, closed, and convex.
\end{assumption}

\begin{assumption}\label{ass:cost}
For every round $t$, the hitting cost $f_t(\cdot):\R^d\mapsto[0,\infty)$ is a finite-valued convex function.
\end{assumption}

\paragraph{Euclidean movement.}
In this paper, we consider two types of movement costs: Euclidean movement $m(\x,\y)=\norm{\x-\y}$ in \Cref{sec:distance} and squared Euclidean movement $m(\x,\y)=\frac12\norm{\x-\y}^2$ in \Cref{sec:upper}.
Two types of movements are commonly used in online optimization and control~\citep{AGG20,GLSW19,ZJLY21}.

\paragraph{Quadratic growth and strong convexity.}
Quadratic growth is a standard relaxation of strong convexity,
because strong convexity with parameter $\lambda$ implies $\lambda$-quadratic growth at a minimizer, while the converse fails~\citep{NNG19,DL18}.
For the Euclidean movement, each hitting cost $f_t(\cdot)$ satisfies the following assumptions:
\begin{assumption}\label{ass:euclidean-qg}
For every $t$, the function $f_t(\cdot)$ attains its global minimum at $\vb_t\in\R^d$ and satisfies $\lambda_t$-quadratic growth defined in \eqref{eq:qg} on $\R^d$ for some $\lambda_t>0$.
\end{assumption}

For the squared Euclidean movement, each hitting cost $f_t(\cdot)$ satisfies the following assumptions.
\begin{assumption}\label{ass:squared-qg}
For every $t$, the function $f_t(\cdot)$ attains its minimum over $\X$ at $\vb_t\in\X$ and satisfies $\lambda$-quadratic growth defined in \eqref{eq:qg} on $\X$ at this minimizer,
in the sense defined in \Cref{sec:results}.
\end{assumption}

\Cref{ass:euclidean-qg} is used for \Cref{sec:distance} (standard Euclidean movement), while \Cref{ass:squared-qg} is used for \Cref{sec:upper} (squared Euclidean movement).
In both cases, quadratic growth is stated relative to a fixed minimizer and therefore implies uniqueness of that minimizer.

\paragraph{Smoothness and a gradient bound.}
Apart from \Cref{ass:euclidean-qg}, the function $f_t(\cdot)$ also satisfies the following smoothness assumption, which is standard in chasing convex functions~\citep{AGG20}.
\begin{assumption}[$L_t$-Smoothness]\label{ass:euclidean-smooth}
For every $t$, the function $f_t(\cdot)$ is differentiable on $\R^d$,
and there exists $L_t>0$ such that
\[
	\norm{\nabla f_t(\x)-\nabla f_t(\y)}
	\le
	L_t\norm{\x-\y}
	, \text{ for all } \x,\y\in\R^d
. \]
The parameters satisfy $L_t\le\kappa\lambda_t$ for all $t$ and may vary across rounds,
where $\lambda$ is the quadratic growth parameter in \Cref{ass:euclidean-qg} and $\kappa$ is a fixed condition number.
The condition of $L_t$-smoothness and $\lambda_t$ quadratic gworth implies that $\kappa\ge 1$.
\end{assumption}
The following gradient bound is a standard consequence of the descent inequality~\citep[Lemma~3.4]{Bub15}.
\begin{lemma}\label{lem:gradient}
Let $f(\cdot):\R^d\mapsto\R$ be $L$-smooth for some constant $L>0$,
and suppose that $f(\cdot)$ attains its global minimum value $f^\star$.
Then, for every $\x\in\R^d$,
\[
	\norm{\nabla f(\x)}^2
	\le 2L(f(\x)-f^\star)
. \]
\end{lemma}

\paragraph{Potential method.}
Fix a comparator sequence $\{\ub\}_{t=0}^T$ with $\ub_0=\x_0$.
Following the standard potential argument~\citep{AGG20},
for either movement cost it suffices to construct a nonnegative potential $\Phi_t$ with $\Phi_0=0$ such that
\[
	f_t(\x_t)+m(\x_t,\x_{t-1})+\Phi_t-\Phi_{t-1}
	\le
	C(f_t(\ub_t)+m(\ub_t,\ub_{t-1}))
, \]
for every round and every comparator sequence.
Therefore, we can prove the $C$-competitiveness by summing over $t=1,\dots,T$:
\[
	\sum_{t=0}^T f_t(\x_t) + m(\x_t,\x_{t-1})
	\le
	\sum_{t=0}^T f_t(\x_t) + m(\x_t,\x_{t-1}) + \Phi_T - \Phi_0
	\le
	C\left(\sum_{t=0}^T f_t(\ub_t) + m(\ub_t,\ub_{t-1})\right)
. \]

\paragraph{Randomization.}
For either movement cost, the one-round cost is jointly convex in the current and previous decisions.
Consequently, randomization can not improve the optimal competitive ratio against oblivious inputs, as observed in \citet{BGKPSS15,AGG20}.

\section{Euclidean movement}\label{sec:distance}
This section presents our proposed method and proves \Cref{thm:distance} for instances in $\Fk$.
The algorithm combines a balanced proximal step with a clipped correction and is summarized in \Cref{alg:cpd}.

\subsection{Clipped Proximal Descent}
\begin{algorithm}[t]
	\caption{Clipped proximal descent}\label{alg:cpd}
	\KwInput{feasible set $\X$, initial decision $\x_0\in\X$, and $\kappa\ge1$, and parameters $r,\beta$ from \eqref{eq:params}.}
	\For{$t=1,\dots,T$}{
		Observe $f_t(\cdot)$ and compute $\vb_t$\;
		$\displaystyle\vb_t^\X=\Pi_{\argmin_{\z\in\X}g_t(\z)}\{\x_{t-1}\}$\;
		\eIf{
			\nllabel{line:if}
			$g_t(\vb_t^\X) \ge \beta\norm{\vb_t^\X-\x_{t-1}}$
		}{
			$\x_t=\vb_t^\X$\;
		}{
			Choose the smallest $\mu_t>0$ satisfying $g_t(\y_t(\mu_t))=\beta\norm{\y_t(\mu_t)-\x_{t-1}}$ \nllabel{line:prox},
			$\y_t=\y_t(\mu_t)$,
			$s_t=\norm{\y_t-\x_{t-1}}$,
			$\displaystyle\gamma_t=\frac{r^2s_t}{s_t+\beta\mu_t}$,
			$\displaystyle\eta_t=\min\left\{\gamma_t,\frac{rs_t}{\norm{\y_t-\vb_t}}\right\}$,
			$\displaystyle\x_t=\Pi_{\X\cap\{\z:g_t(\z)\le g_t(\y_t)\}} \left\{(1-\eta_t)\y_t+\eta_t\vb_t\right\}$
			\nllabel{line:projection}\;
		}
		Play $\x_t$ and incur $f_t(\x_t)+\norm{\x_t-\x_{t-1}}$\;
	}
\end{algorithm}

At each round $t$, the algorithm starts from $\x_{t-1}\in\X$ and observes the hitting cost $f_t(\cdot)$. It computes the global minimizer $\vb_t$ from \Cref{ass:euclidean-qg} and defines $g_t(\z)=f_t(\z)-f_t(\vb_t)$.
This standard normalization~\cite{AGG20} preserves the growth and smoothness parameters,
so $g_t(\vb_t)=0$ and $L_t\le\kappa\lambda_t$.
By \Cref{ass:euclidean-qg,ass:euclidean-smooth} and \Cref{lem:gradient},
for every $\z\in\R^d$ we have
\[
g_t(\z)\ge\frac{\lambda_t}{2}\norm{\z-\vb_t}^2,
\qquad
\norm{\nabla g_t(\z)}^2\le2L_tg_t(\z).
\]
Then we show that a competitive guarantee for these shifted costs also holds for the hitting costs $f_t(\cdot)$.

For the update, we set
\begin{equation}\label{eq:params}
r=\frac{1}{2\kappa^{1/3}},\qquad
\beta=1+3r,\qquad
C=\frac{32}{7}\bigl(1+\kappa^{1/3}\bigr)
 =\frac{8(2+4r)}{7r},
\end{equation}
where $0<r\le\frac12$ and $\kappa r^3=\frac18$.

We next find the constrained minimizer $\vb_t^\X$ closest to $\x_{t-1}$.
If $g_t(\vb_t^\X)\ge\beta\norm{\vb_t^\X-\x_{t-1}}$, it sets $\x_t=\vb_t^\X$.  Every feasible comparator pays a hitting cost of at least $g_t(\vb_t^\X)$, and the condition
bounds the movement by $g_t(\vb_t^\X)/\beta$.

Otherwise, the algorithm enters the balanced case. For $\mu>0$, consider the proximal point
\begin{equation*}
\y_t(\mu)=\argmin_{\z\in\X} \left\{g_t(\z) +\frac{\norm{\z-\x_{t-1}}^2}{2\mu}\right\}.
\end{equation*}
The quadratic term penalizes distance from the previous decision and $\mu$ controls its weight relative to the hitting cost.
Choose the smallest $\mu_t>0$ such that $g_t(\y_t(\mu_t)) =\beta\norm{\y_t(\mu_t)-\x_{t-1}}$, and set $\y_t=\y_t(\mu_t)$ and $s_t=\norm{\y_t-\x_{t-1}}$. Thus $g_t(\y_t)=\beta s_t$: the hitting cost at the proximal point equals $\beta$ times the Euclidean movement needed to reach it. 

Next, take a clipped step from $\y_t$ toward the global minimizer $\vb_t$, using the fraction $\eta_t$ defined by
\[
\gamma_t=\frac{r^2s_t}{s_t+\beta\mu_t},
\qquad
\eta_t=\min\left\{\gamma_t,
\frac{rs_t}{\norm{\y_t-\vb_t}}\right\}.
\]
The resulting point is $(1-\eta_t)\y_t+\eta_t\vb_t$.
The cap $\eta_t\le\gamma_t$ controls the localization error in the
analysis, while $\eta_t\norm{\y_t-\vb_t}\le rs_t$
limits the additional movement to a fraction $r$ of the proximal step.

\begin{figure}[t]
\centering
\includegraphics{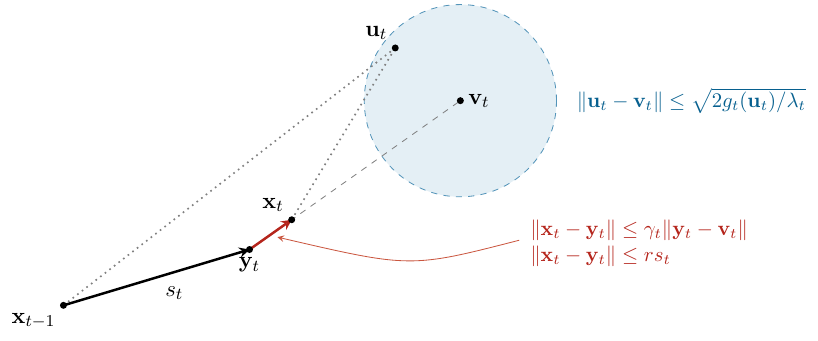}
\caption{A balanced round of \Cref{alg:cpd} with $\X=\R^d$.
The proximal step moves $\x_{t-1}$ to $\y_t$, and the clipped correction moves $\y_t$ toward $\vb_t$.}
\label{fig:correction}
\end{figure}

Finally, project this point onto $\X\cap\{\z:g_t(\z)\le g_t(\y_t)\}$
to obtain $\x_t$. The projection ensures feasibility and $g_t(\x_t)\le g_t(\y_t)$. If $\X=\R^d$, \Cref{ass:cost} gives
\[
g_t\bigl((1-\eta_t)\y_t+\eta_t\vb_t\bigr)
\le(1-\eta_t)g_t(\y_t)+\eta_tg_t(\vb_t)
\le g_t(\y_t),
\]
so the projection leaves the corrected point unchanged, as illustrated in \Cref{fig:correction}.
The complete update is summarized in \Cref{alg:cpd}.
It uses exact minimization, proximal, projection oracles and a one-dimensional search for $\mu_t$
Its inputs are $\kappa,\x_0,\X$, and $r,\beta$ can be computed using $\kappa$.

\subsection{Theoretical analysis}
Firstly we demonstrate that the algorithm is well defined and the parameters are bounded using the following lemma.
\begin{lemma}\label{lem:welldefined}
In the case of line 7 in \Cref{alg:cpd}, the smallest positive balancing parameter $\mu_t$ exists,
$s_t>0$, and $g_t(\y_t)=\beta s_t>0$.
Moreover, $\y_t\ne\vb_t$ and $0<\eta_t\le\gamma_t<r^2\le\frac14$.
\end{lemma}
The proof of \Cref{lem:welldefined} is deferred to \Cref{app:welldefined}.
After the existence of $\mu_t$ is established, one-dimensional bisection can be used to find it efficiently in its implementation.
The remaining parameters can be computed in closed form, and $\x_t$ can be computed using a projection oracle.
Therefore, every update of \Cref{alg:cpd} is well defined.
\par 
We next show that the competitive ratio of \Cref{alg:cpd} is $C$ in \eqref{eq:params}, which proves \Cref{thm:distance}.
\begin{proof}
\Cref{alg:cpd} evaluates $g_t(\z)=f_t(\z)-f_t(\vb_t)$ instead of $f_t(\z)$, which is a standard normalization~\citep{AGG20}.
We first show that this normalization preserves the competitive ratio.
For any comparator sequence $\{\ub_t\}_{t=0}^T$ with $\ub_0=\x_0$,
if a online algorithm achieves a $C$-competitive guarantee on the shifted costs $g_t$, namely
\[
	\sum_{t=1}^T\bigl(g_t(\x_t)+\norm{\x_t-\x_{t-1}}\bigr)
	\le C\sum_{t=1}^T\bigl(g_t(\ub_t)+\norm{\ub_t-\ub_{t-1}}\bigr)
, \]
then it also achieves a $C$-competitive guarantee on the original costs $f_t(\cdot)$, since
\begin{align*}
	&~
	\sum_{t=1}^T \left(f_t(\x_t) + \|\x_t-\x_{t-1}\|\right)
	=
	\sum_{t=1}^T \left(g_t(\x_t) + \|\x_t-\x_{t-1}\|\right)
	+ \sum_{t=1}^T f_t(\vb_t)
	\\
	\le&~
	C \sum_{t=1}^T \left(g_t(\ub_t) + \|\ub_t-\ub_{t-1}\|\right)
	+ \sum_{t=1}^T f_t(\vb_t)
	\le
	C \sum_{t=1}^T \left(f_t(\ub_t) + \|\ub_t-\ub_{t-1}\|\right)
\end{align*}
where the last inequality uses $C\ge1$ and $f_t(\vb_t)\ge0$ from \Cref{ass:cost}.
Therefore, it suffices to analyse the competitive ratio for the shifted costs $g_t(\cdot)$.
We first establish the amortized bound
\begin{equation}\label{eq:amortized-cost}
	g_t(\x_t) + \norm{\x_t-\x_{t-1}}
	+ C\left(\norm{\x_t-\ub_t}-\norm{\x_{t-1}-\ub_t}\right)
	\le Cg_t(\ub_t)
. \end{equation}
Suppose first that the condition in line 4 holds.
Then $\x_t=\vb_t^\X$ and
\[
	\norm{\vb_{t,\X}-\x_{t-1}}
	\le
	\frac{g_t(\vb_t^\X)}{\beta}
. \]
The triangle inequality gives
\begin{align*}
	&~
	g_t(\x_t) + \norm{\x_t-\x_{t-1}} + C\left(\norm{\x_t-\ub_t}-\norm{\x_{t-1}-\ub_t}\right)
	\\
	\le
	&~
	g_t(\vb_t^\X)+(1+C)\norm{\vb_t^\X-\x_{t-1}}
	\le
	\left(1+\frac{1+C}{\beta}\right) g_t(\vb_t^\X)
	\le
	C g_t(\vb_t^\X)
. \end{align*}
The last inequality is by $C=\frac{32}{7}(1+\kappa^{1/3})$, and
\[
	C-1-(1+C)\beta
	=
	\frac{3\left(\frac{68}{21}\kappa^{1/3}+\frac{25}{7}\right)}
	{2\kappa^{1/3}+3}
	\ge
	0.
\]
Since $\vb_t^\X$ minimizes $g_t$ over $\X$,
the preceding bound is at most $Cg_t(\vb_t^\X)\le Cg_t(\ub_t)$.
This proves \eqref{eq:amortized-cost} in the first case.
\par 
Suppose that the condition in line 4 of \Cref{alg:cpd} is false.
The projection in line 7 ensures $g_t(\x_t)\le g_t(\y_t)$.
Its projection set contains $\y_t$, so the expansion gives
\[
	\norm{\x_t-\y_t}
	\le \norm{(1-\eta_t)\y_t+\eta_t\vb_t-\y_t}
	=\eta_t\norm{\y_t-\vb_t}
	\le r \norm{\y_t-\x_{t-1}}
. \]
Combining this estimate with the triangle inequality and $g_t(\y_t)=\beta\norm{\y_t-\x_{t-1}}$, we obtain
\begin{align*}
	g_t(\x_t)+\norm{\x_t-\x_{t-1}}
	&\le g_t(\y_t)+\norm{\y_t-\x_{t-1}}+\norm{\x_t-\y_t}
	\\
	&\le \left(2+4r\right) \norm{\y_t-\x_{t-1}}
	= C\frac{7}{16\kappa^{1/3}}\norm{\y_t-\x_{t-1}}.
\end{align*}
We next bound this term use the following lemma:
\begin{lemma}\label{lem:progress}
Suppose that the condition in Line 4 of Algorithm~\ref{alg:cpd} is false.
Then every $\ub_t\in\X$ satisfies
\[
	\norm{\x_{t-1}-\ub_t}-\norm{\x_t-\ub_t}+g_t(\ub_t)
	\ge \frac{7}{16\kappa^{1/3}}\norm{\y_t-\x_{t-1}}
. \]
\end{lemma}
Combining this lemma with the preceding cost estimate yields
\[
	g_t(\x_t)+\norm{\x_t-\x_{t-1}}
	\le C\left(
	\norm{\x_{t-1}-\ub_t}-\norm{\x_t-\ub_t}+g_t(\ub_t) \right)
, \]
which proves \eqref{eq:amortized-cost} in the second case.
\par 
To sum the amortized bounds, apply the triangle inequality
\[
	\norm{\x_{t-1}-\ub_t}
	\le \norm{\x_{t-1}-\ub_{t-1}}+\norm{\ub_t-\ub_{t-1}}
. \]
Substituting into \eqref{eq:amortized-cost} gives
\[
	g_t(\x_t)+\norm{\x_t-\x_{t-1}} +C\left(\norm{\x_t-\ub_t}-\norm{\x_{t-1}-\ub_{t-1}}\right)
	\le C\left(g_t(\ub_t)+\norm{\ub_t-\ub_{t-1}}\right)
. \]
Summing over $t=1,\dots,T$ and using $\ub_0=\x_0$, we obtain
\[
	\sum_{t=1}^T
	\left(g_t(\x_t)+\norm{\x_t-\x_{t-1}}\right)
	\le
	C\sum_{t=1}^T
	\left(g_t(\ub_t)+\norm{\ub_t-\ub_{t-1}}\right)
. \]
This concludes the proof of the $C$-competitive guarantee for \Cref{alg:cpd}.
\end{proof}
The strongly convex construction in \cite[Proposition~13]{AGG20} gives the matching $\Omega(\kappa^{1/3})$ lower bound.
Its independent random signs also establish the expected lower bound against randomized algorithms, completing the minimax claim.

\section{Squared movement}\label{sec:upper}
This section establishes the upper and lower bounds in \Cref{thm:squared} by the R-OBD~\citep{GLSW19} algorithm.
Throughout, we consider instances in $\Ql$, with minimizers $\vb_t$ specified in \Cref{ass:squared-qg}.

\subsection{Regularized online balanced descent}
Given the quadratic growth parameter $\lambda$ and the feasible set $\X$, choose $q>0$ satisfying
\[
	\min_{0\le d<1} \Lambda(q,d) = \lambda
, \]
where $\Lambda$ is defined in \eqref{eq:squared:Lambda:phi}.
Choose $\delta=\argmin_{0\le d<1}\Lambda(q,d)$ and set the weights $a,b>0$ according to \eqref{eq:weights}.
At each round $t$, R-OBD updates
\[
	\x_t
	=
	\argmin_{\x\in\X}
	\left\{
		f_t(\x)
		+\frac{a}{2}\|\x-\x_{t-1}\|^2
		+\frac{b}{2}\|\x-\vb_t\|^2
	\right\}
, \]
where $\vb_t$ minimizes $f_t(\cdot)$ over $\X$ and $\x_{t-1}$ is the previous decision.
\par 
By completing the square norms, R-OBD can be viewed as the equivalent proximal update 
\begin{equation*}
\x_t = \argmin_{\x\in\X} \{f_t(\x)+(a+b)/2\norm{\x-\cb_t}^2 \},\quad  \cb_t=(a\x_{t-1}+b\vb_t)/(a+b).
\end{equation*}
The center $\cb_t$ is a convex combination of $\x_{t-1}$ and $\vb_t$.
Our analysis requires only convexity and quadratic growth of $f_t(\cdot)$.
In particular, the bound $f_t(\vb_t)-f_t(\ub) \le -\lambda/2\norm{\ub-\vb_t}^2$ is used to establish the competitive guarantee below.

\begin{lemma}\label{lem:certificate}
Let $q>0$ and $0\le\delta<1$, and set the parameters of R-OBD as
\begin{equation}\label{eq:weights}
	h=1+q\delta^2,
	\qquad
	a=\frac{h}{1-\delta},
	\qquad
	b=\frac{a^2}{q(1+\delta)}.
\end{equation}
If $\lambda\ge\Lambda(q,\delta)$, then R-OBD with weights $a$ and $b$ is $(1+q)$-competitive on $\Ql$.
\end{lemma}
\begin{proof}
Let $c=Ca/(q(1+\delta))$.
Fix any comparator sequence $\{\ub_t\}_{t=0}^T$ with $\ub_0=\x_0$,
and define
\[
	\Phi_t=\frac{c}{2}\norm{\x_t-\ub_t}^2
. \]
We first establish the amortized bound
\begin{equation}\label{eq:rob-d-amortized}
	f_t(\x_t)+\frac12\norm{\x_t-\x_{t-1}}^2
	+\Phi_t-\Phi_{t-1}
	\le
	C\left(f_t(\ub_t)+\frac12\norm{\ub_t-\ub_{t-1}}^2\right)
. \end{equation}
\par 
In round $t$, optimality of $\x_t$ imply that, for every $\y\in\X$,
\[
	f_t(\x_t) + \frac{a}{2} \|\x_t-\x_{t-1}\|^2 + \frac{b}{2} \|\x_t-\vb_t\|^2
	\le
	f_t(\y_\varepsilon) + \frac{a}{2} \|\y_\varepsilon-\x_{t-1}\|^2 + \frac{b}{2} \|\y_\varepsilon-\vb_t\|^2
, \]
where $\y_\varepsilon=\x_t+\varepsilon(\y-\x_t)$.
Rearranging and expanding the quadratic terms gives
\[
	f_t(\x_t) - f_t(\y)
	\le
	\varepsilon a \left< \x_t - \x_{t-1}, \y - \x_t \right>
	+ \varepsilon^2 \frac{a}{2} \|\y - \x_t\|^2
	+ \varepsilon b \left< \x_t - \vb_t, \y - \x_t \right>
	+ \varepsilon^2 \frac{b}{2} \|\y - \x_t\|^2
. \]
And by the convexity of $f_t(\cdot)$, we have
\[
	\varepsilon(f_t(\x_t) - f_t(\y))
	\le
	f_t(\x_t) - f_t(\y)
	\le
	\varepsilon \left< a(\x_t-\x_{t-1}) + b(\x_t-\vb_t), \y - \x_t \right>
	+ \varepsilon^2 \frac{a+b}{2} \|\y - \x_t\|^2
. \]
This inequality holds for every $\varepsilon\in(0,1)$, and dividing by $\varepsilon$ and letting $\varepsilon\to0$ gives
\[
	f_t(\x_t)-f_t(\y)
	\le
	\left< a(\x_t-\x_{t-1}) + b(\x_t-\vb_t), \y - \x_t \right>
. \]
\par 
Apply the above inequality at $\y=\ub_t$ and $\y=\vb_t$,
with weights $1-\delta$ and $\delta$ respectively.
Then by summing the two inequalities, we obtain
\[
	f_t(\x_t)
	\le
	(1-\delta)f_t(\ub_t)+\delta f_t(\vb_t)
	- \left< a(\x_t-\x_{t-1}) + b(\x_t-\vb_t), \x_t - \vb_t - (1-\delta)(\ub_t-\vb_t) \right>
. \]
By \Cref{ass:squared-qg},
\[
	f_t(\vb_t)-f_t(\ub_t)
	\le
	-\frac{\lambda}{2}\norm{\ub_t-\vb_t}^2
	, \quad
	-qf_t(\ub_t)
	\le
	-q\frac{\lambda}{2}\norm{\ub_t-\vb_t}^2 - qf_t(\vb_t)
. \]
Terms $(1-\delta)f_t(\ub_t)+\delta f_t(\vb_t)=f_t(\ub_t) + \delta(f_t(\vb_t)-f_t(\ub_t))$ are bounded by the first above inequality,
and adding the second above inequality and the preceding inequalities yields
\begin{equation}\label{eq:twocomp}
\begin{aligned}
	f_t(\x_t)-Cf_t(\ub_t)
	\le&~
	- (q+\delta)\frac{\lambda}{2}\|\ub_t-\vb_t\|^2 -qf_t(\vb_t)
	\\&
	- \left< a(\x_t-\x_{t-1})+b(\x_t-\vb_t), \x_t-\vb_t-(1-\delta)(\ub_t-\vb_t) \right>
\end{aligned}
. \end{equation}
Here $C=1+q$ is the competitive ratio.
\par 
We next bound the potential difference.
Expanding the square norm gives
\begin{align*}
	\Phi_{t-1}+\frac{C}{2} \|\ub_t-\ub_{t-1}\|^2
	&=
	\frac{c}{2} \|\x_{t-1}-\ub_{t-1}\|^2 + \frac{C}{2} \|\ub_t-\ub_{t-1}\|^2
	\\
	&=
	\frac{c}{2} \|\x_{t-1}-\ub_t+\ub_t-\ub_{t-1}\|^2 + \frac{C}{2} \|\ub_t-\ub_{t-1}\|^2
	\\
	&=
	\frac{c}{2} \|\x_{t-1}-\ub_t\|^2 + \frac{c+C}{2} \|\ub_t-\ub_{t-1}\|^2
	+ c \left< \x_{t-1}-\ub_t, \ub_t-\ub_{t-1} \right>
	\\
	&=
	\frac{c+C}{2} \|\ub_t-\ub_{t-1} + \frac{c}{C+c} (\x_{t-1} - \ub_t)\|^2
	+ \frac{cC}{2(c+C)} \|\x_{t-1}-\ub_t\|^2
. \end{align*}
Using $cC/(C+c)=h$ gives
\[
	\Phi_{t-1}+\frac{C}{2} \|\ub_t-\ub_{t-1}\|^2
	\ge
	\frac{h}{2} \|\x_{t-1}-\ub_t\|^2
. \]
Consequently,
\[
	\Phi_t-\Phi_{t-1}
	-\frac{C}{2}\|\ub_t-\ub_{t-1}\|^2
	\le
	\frac{c}{2}\|\x_t-\ub_t\|^2
	-\frac{h}{2}\|\x_{t-1}-\ub_t\|^2
. \]
To combine this bound with \eqref{eq:twocomp},
we have
\begin{align*}
	&~
	f_t(\x_t)-C(f_t(\ub_t)+\frac12\|\ub_t-\ub_{t-1}\|^2)
	+\Phi_t-\Phi_{t-1}
	+ \frac{1}{2} \|\x_t-\x_{t-1}\|^2
	\\
	\le&~
	- (q+\delta)\frac{\lambda}{2}\|\ub_t-\vb_t\|^2 -qf_t(\vb_t)
	- \left< a(\x_t-\x_{t-1})+b(\x_t-\vb_t), \x_t-\vb_t-(1-\delta)(\ub_t-\vb_t) \right>
	\\&
	+ \frac{c}{2}\|\x_t-\ub_t\|^2
	- \frac{h}{2}\|\x_{t-1}-\ub_t\|^2
	+ \frac{1}{2} \|\x_t-\x_{t-1}\|^2
. \end{align*}
Expand the squared norms terms by
\[
	\x_t-\ub_t
	=
	(\x_t-\vb_t)-(\ub_t-\vb_t)
	, \quad
	\x_{t-1}-\ub_t
	=
	(\x_t-\vb_t)-(\ub_t-\vb_t)-(\x_t-\x_{t-1})
. \]
And inner product terms involving $\ub_t-\vb_t$ are cancelled by $(1-\delta)a=h$ and $(1-\delta)b=c-h$.
Thus,
\begin{align*}
	&~
	f_t(\x_t) + \frac{1}{2} \|\x_t-\x_{t-1}\|^2
	- C(f_t(\ub_t) + \frac12\|\ub_t-\ub_{t-1}\|^2)
	+ \Phi_t-\Phi_{t-1}
	\\
	\le&~
	-\frac{(1+\delta)b}{2}\|\x_t-\vb_t\|^2
	-\delta a \left<\x_t-\vb_t, \x_t-\x_{t-1}\right>
	- \frac{q\delta^2}{2} \|\x_t-\x_{t-1}\|^2
	- qf_t(\vb_t)
	\\ &~
	+ \frac{(1-\delta)b-(q+\delta)\lambda}{2} \|\ub_t-\vb_t\|^2
. \end{align*}
Completing the first three terms in the right-hand side by substituting $(1+\delta)b=a^2/q$ and $\|A+B\|^2=\|A\|+2\left<A,B\right>+\|B\|^2$,
and substituting $(1-\delta)b=(q+\delta)\Lambda(q,\delta)$ give
\begin{align*}
	&~
	f_t(\x_t) + \frac{1}{2} \|\x_t-\x_{t-1}\|^2
	- C(f_t(\ub_t) + \frac12\|\ub_t-\ub_{t-1}\|^2)
	+ \Phi_t-\Phi_{t-1}
	\\
	\le&~
	- \frac{1}{2q} \|a(\x_t-\vb_t)+q\delta(\x_t-\x_{t-1})\|^2
	- q f_t(\vb_t)
	- \frac{q+\delta}{2}(\lambda-\Lambda(q,\delta)) \|\ub_t-\vb_t\|^2
	\le
	0
. \end{align*}
Here the last inequality uses \Cref{ass:cost} $f_t(\cdot)\ge0$ and the pre-request $\lambda\ge\Lambda(q,\delta)$,
which proves \eqref{eq:rob-d-amortized}.
\par 
Summing \eqref{eq:rob-d-amortized} over $t=1,\dots,T$,
and using $\Phi_0=0$ and $\Phi_T\ge0$, we obtain
\begin{align*}
	\sum_{t=1}^T \left(f_t(\x_t)+\frac12\norm{\x_t-\x_{t-1}}^2\right)
	&\le
	\sum_{t=1}^T \left(f_t(\x_t)+\frac12\norm{\x_t-\x_{t-1}}^2\right)
	+ \Phi_T - 0
	\\
	&\le
	C\sum_{t=1}^T \left(f_t(\ub_t)+\frac12\norm{\ub_t-\ub_{t-1}}^2\right)
. \end{align*}
This concludes the proof of the $C$-competitive guarantee.
\end{proof}

The best ratio provided by \Cref{lem:certificate} is the smallest $1+q$ for which some $\delta$ satisfies $\Lambda(q,\delta)\le\lambda$. The following lemma characterizes this.
\begin{lemma}\label{lem:scalar}
For every $q>0$, the function $\Lambda(q,\cdot)$ has a unique minimizer $\ds(q)$ on $[0,1)$, and $\ds(q)$ is the unique root in $(0,1/\sqrt3)$ of
\[
	P_q(\delta)=2\delta(2-\delta^2)q^2+\delta(2+3\delta-\delta^3)q+3\delta^2-1
. \]
Moreover, $P_q<0$ on $[0,\ds(q))$ and $P_q>0$ on $(\ds(q),1)$. The function $\phi$ defined in \eqref{eq:squared:Lambda:phi} equals $\Lambda(q,\ds(q))$, is continuous and strictly decreasing on $(0,\infty)$, and satisfies
\[
	\frac{1}{q(q+1)}\le\phi(q)\le\frac{1}{q^2}
. \]
\end{lemma}
By the last inequalities, the inverse $\phi^{-1}(\lambda)$ exists for every $\lambda>0$ and is the unique $q$ satisfying $\min_{0\le d<1}\Lambda(q,d)=\lambda$.

\subsection{Lower bound}\label{sec:lower}
This section proves the lower bound of \Cref{thm:squared}.
The construction realizes the equality conditions of \Cref{lem:certificate}.
Every online algorithm pays at least the cost of a reference path in expectation, and an offline comparator pays a fraction $1/(1+\qs(\lambda))$ of this cost per round.
Throughout the section, $m(\x,\y)=\frac12\norm{\x-\y}^2$, $\X=\R^d$, and $\x_0=\0$.

\begin{lemma}[Lower bound for squared Euclidean movement]
\label{lem:lower}
Fix $\lambda>0$ and an infinite-dimensional real Hilbert space
$\mathcal H_\infty$.
Let $\D=\{\R^d:d\in\mathbb N,d\ge1\}\cup\{\H_\infty\}$ and let $\A$ be the class of all randomized online algorithms
defined on the spaces in $\D$.
For each $\H\in\D$, let $\F_\lambda(\H)$ be the class of nonnegative convex functions on $\mathcal H$ satisfying the $\lambda$-quadratic growth condition.
Then the dimension-uniform minimax competitive ratio satisfies
\[
	\inf_{A\in\A}
	\sup_{\substack{
		T\in\mathbb N,\ \H\in\D\\
		(f_1,\dots,f_T)\in\F_\lambda(\mathcal H)^T\\
	}}
	\frac{\E_\xi[\ALG_T]}{\OPT_T}
	\ge 1+q_*(\lambda)
. \]
Here the expectation is taken over the algorithm's internal randomness $\xi$.
\end{lemma}
Combining \Cref{lem:certificate} and \Cref{lem:lower} proves \Cref{thm:squared}.
The proof is deferred to \Cref{app:lower}.
In other words, for every $\lambda>0$ and every $R<1+q_*(\lambda)$,
any randomized online algorithm defined across all finite-dimensional Euclidean spaces and Hilbert space,
a finite horizon $T$, and a sequence of cost functions,
we have $\E_{\xi}[\ALG_T] > R\,\OPT_T,$ where $\xi$ denotes the algorithm's internal randomness.
Consequently, no such algorithm can achieve a competitive ratio strictly smaller than $1+q_*(\lambda)$ uniformly over all finite dimensions.

Here, other algorithms may depend on the dimension and the announced horizon.
However, we use dimension-uniformity to refers to a competitive guarantee whose ratio is independent of the dimension $d$,
and the corresponding algorithm works for any finite dimension $d$ and any horizon $T$, but and its parameters are independent of the dimension $d$.
The same lower bound also holds on any fixed infinite-dimensional real Hilbert space.
\Cref{lem:lower} establishes the lower bound for the dimension-uniform minimax competitive ratio of any randomized online algorithm on the class of convex functions with $\lambda$-quadratic growth and squared Euclidean movement cost.

\section{Conclusion}\label{sec:conclusion}
In this paper, we investigate convex function chasing with quadratic growth under two movement costs. For Euclidean movement, we propose Clipped Proximal Descent (CPD), which achieves an $O(\kappa^{1/3})$ competitive ratio for smooth convex functions satisfying quadratic growth. This matches the existing lower bound~\citep{AGG20} and resolves the gap for $\kappa$-well-conditioned functions under a weaker assumption than strong convexity. For squared Euclidean movement, we establish the exact optimal dimension-free competitive ratio for convex functions with quadratic growth through a tight analysis of R-OBD and a matching lower bound against randomized algorithms.

\section*{Disclosure of generative AI use}
\phantomsection
\addcontentsline{toc}{section}{Disclosure of generative AI use}
We use GPT-6 Astra to assist with exploring proof strategies and checking mathematical derivations.
The authors also use  GPT-6 Astra for identifying relevant literature and polishing the paper.
The tool materially affected \Cref{sec:distance} and \Cref{sec:upper}. The authors verified the correctness and originality of all content, including references.

\bibliography{refs}

\newpage
\appendix
\section{Proofs of supporting results}\label{app:supporting}

%

\subsection{Proof of \texorpdfstring{\Cref{lem:welldefined}}{Lemma \getrefnumber{lem:welldefined}}}\label{app:welldefined}
\begin{proof}
Fix a round $t$. \Cref{ass:euclidean-qg} gives
$g_t(\mathbf{z})\ge\lambda_t\norm{\mathbf{z}-\mathbf{v}_t}^2/2
\to\infty$ as $\norm{\mathbf{z}}\to\infty$.
Thus $g_t$ is coercive. By \Cref{ass:cost}, it is also continuous
and convex. Together with \Cref{ass:domain}, these properties imply
that its minimizers over $\X$ form a nonempty compact convex set.
The projection defining $\mathbf{v}_t^\X$ is therefore unique. For $\mu>0$, recall that
\begin{equation}\label{eq:prox-path}
\mathbf{y}_t(\mu)=\argmin_{\mathbf{z}\in\X}
\left\{g_t(\mathbf{z})
+\frac{\norm{\mathbf{z}-\mathbf{x}_{t-1}}^2}{2\mu}\right\}.
\end{equation} 
The objective in \eqref{eq:prox-path} is coercive
and $(1/\mu)$-strongly convex, so $\mathbf{y}_t(\mu)$ is also unique.
Comparison with $\mathbf{v}_t^\X$ gives
\begin{align*}
g_t(\mathbf{y}_t(\mu))
 +\frac{\norm{\mathbf{y}_t(\mu)-\mathbf{x}_{t-1}}^2}{2\mu}
&\le g_t(\mathbf{v}_t^\X)
 +\frac{\norm{\mathbf{v}_t^\X-\mathbf{x}_{t-1}}^2}{2\mu}.
\end{align*}
Using $g_t(\mathbf{y}_t(\mu))\ge g_t(\mathbf{v}_t^\X)$ and dropping
nonnegative terms yields
\begin{equation}\label{eq:prox-path-bounds}
\begin{aligned}
\norm{\mathbf{y}_t(\mu)-\mathbf{x}_{t-1}}
&\le\norm{\mathbf{v}_t^\X-\mathbf{x}_{t-1}},\\
0\le g_t(\mathbf{y}_t(\mu))-g_t(\mathbf{v}_t^\X)
&\le\frac{\norm{\mathbf{v}_t^\X-\mathbf{x}_{t-1}}^2}{2\mu}.
\end{aligned}
\end{equation}

To show continuity of the proximal path, take $\mu,\nu>0$.
The two optimality conditions are
\begin{align*}
\left\langle\nabla g_t(\mathbf{y}_t(\mu))
 +\frac{\mathbf{y}_t(\mu)-\mathbf{x}_{t-1}}{\mu},
 \mathbf{y}_t(\nu)-\mathbf{y}_t(\mu)\right\rangle&\ge0,\\
\left\langle\nabla g_t(\mathbf{y}_t(\nu))
 +\frac{\mathbf{y}_t(\nu)-\mathbf{x}_{t-1}}{\nu},
 \mathbf{y}_t(\mu)-\mathbf{y}_t(\nu)\right\rangle&\ge0.
\end{align*}
\Cref{ass:cost} gives the two inequalities
\begin{align*}
g_t(\mathbf{y}_t(\mu))-g_t(\mathbf{y}_t(\nu))
&\ge\ip{\nabla g_t(\mathbf{y}_t(\nu))}
        {\mathbf{y}_t(\mu)-\mathbf{y}_t(\nu)},\\
g_t(\mathbf{y}_t(\nu))-g_t(\mathbf{y}_t(\mu))
&\ge\ip{\nabla g_t(\mathbf{y}_t(\mu))}
        {\mathbf{y}_t(\nu)-\mathbf{y}_t(\mu)}.
\end{align*}
Adding them yields
\[
\left\langle\nabla g_t(\mathbf{y}_t(\mu))
-\nabla g_t(\mathbf{y}_t(\nu)),
\mathbf{y}_t(\mu)-\mathbf{y}_t(\nu)\right\rangle\ge0.
\]
Adding the optimality conditions and using this inequality, we obtain
\begin{align*}
\frac{\norm{\mathbf{y}_t(\mu)-\mathbf{y}_t(\nu)}^2}{\mu}
&\le\left(\frac1\nu-\frac1\mu\right)
  \ip{\mathbf{y}_t(\nu)-\mathbf{x}_{t-1}}
     {\mathbf{y}_t(\mu)-\mathbf{y}_t(\nu)}\\
&\le\frac{|\mu-\nu|}{\mu\nu}
  \norm{\mathbf{v}_t^\X-\mathbf{x}_{t-1}}
  \norm{\mathbf{y}_t(\mu)-\mathbf{y}_t(\nu)},
\end{align*}
where the last step uses Cauchy--Schwarz and
\eqref{eq:prox-path-bounds}. Dividing when the distance is nonzero
(the zero case is immediate) gives
\[
\norm{\mathbf{y}_t(\mu)-\mathbf{y}_t(\nu)}
\le\frac{|\mu-\nu|}{\nu}\norm{\mathbf{v}_t^\X-\mathbf{x}_{t-1}}
\longrightarrow0\quad\text{as }\mu\to\nu.
\]

Comparing the proximal objective with its value at
$\mathbf{x}_{t-1}$ gives
\begin{align*}
g_t(\mathbf{y}_t(\mu))
 +\frac{\norm{\mathbf{y}_t(\mu)-\mathbf{x}_{t-1}}^2}{2\mu}
&\le g_t(\mathbf{x}_{t-1}),\\
\norm{\mathbf{y}_t(\mu)-\mathbf{x}_{t-1}}^2
&\le2\mu g_t(\mathbf{x}_{t-1})\longrightarrow0
\quad(\mu\downarrow0).
\end{align*}
As $\mu\to\infty$, \eqref{eq:prox-path-bounds} makes the proximal
points bounded. Any convergent subsequence
$\mathbf{y}_t(\mu_j)\to\bar{\mathbf{y}}_t$ with $\mu_j\to\infty$
satisfies
\[
\bar{\mathbf{y}}_t\in\X,\qquad
g_t(\bar{\mathbf{y}}_t)=g_t(\mathbf{v}_t^\X),\qquad
\norm{\bar{\mathbf{y}}_t-\mathbf{x}_{t-1}}
\le\norm{\mathbf{v}_t^\X-\mathbf{x}_{t-1}}.
\]
The closest constrained minimizer is unique, so every such limit is
$\mathbf{v}_t^\X$. We have therefore shown that
$\mathbf{y}_t(\mu)\to\mathbf{x}_{t-1}$ as $\mu\downarrow0$ and
$\mathbf{y}_t(\mu)\to\mathbf{v}_t^\X$ as $\mu\to\infty$.

In the balanced case,
$g_t(\mathbf{v}_t^\X)<\beta\norm{\mathbf{v}_t^\X-\mathbf{x}_{t-1}}$.
Thus $\mathbf{x}_{t-1}$ is not a constrained minimizer: otherwise
$\mathbf{v}_t^\X=\mathbf{x}_{t-1}$ would contradict this inequality.
In particular, $g_t(\mathbf{x}_{t-1})>g_t(\mathbf{v}_t^\X)\ge0$.
The limits just proved imply
\begin{align*}
\lim_{\mu\downarrow0}
\bigl(g_t(\mathbf{y}_t(\mu))
 -\beta\norm{\mathbf{y}_t(\mu)-\mathbf{x}_{t-1}}\bigr)
&=g_t(\mathbf{x}_{t-1})>0,\\
\lim_{\mu\to\infty}
\bigl(g_t(\mathbf{y}_t(\mu))
 -\beta\norm{\mathbf{y}_t(\mu)-\mathbf{x}_{t-1}}\bigr)
&=g_t(\mathbf{v}_t^\X)
 -\beta\norm{\mathbf{v}_t^\X-\mathbf{x}_{t-1}}<0.
\end{align*}
Continuity and the intermediate value theorem give a positive root.
The expression is positive for all sufficiently small $\mu>0$, so
the root set is bounded away from zero. Its infimum is a root by
continuity, proving the existence of the smallest root $\mu_t$.

At this root, $\mathbf{y}_t=\mathbf{x}_{t-1}$ would imply
$g_t(\mathbf{x}_{t-1})=0$, contradicting the strict positivity above.
Therefore $s_t>0$ and
$g_t(\mathbf{y}_t)=\beta s_t>0=g_t(\mathbf{v}_t)$, so
$\norm{\mathbf{y}_t-\mathbf{v}_t}>0$.
The definitions give $0<\gamma_t=r^2s_t/(s_t+\beta\mu_t)<r^2\le1/4$
and $0<\eta_t\le\gamma_t$.
Finally, \Cref{ass:domain,ass:cost} imply that the projection set
is closed and convex. It contains $\mathbf{y}_t$, so the final
projection exists and is unique.
\end{proof}

\subsection{Proof of \texorpdfstring{\Cref{lem:progress}}{Lemma \getrefnumber{lem:progress}}}\label{app:progress}
\begin{proof}
Fix $t$ and write $g=g_t$, $p=x_{t-1}$, $x=x_t$, $u=u_t$, $v=v_t$, $L=L_t$, and $\lambda=\lambda_t$.
Suppose that the condition in line 4 is false,
and retain the notation of the theorem proof.
For every $u\in\X$,
by the proximal optimality condition and convexity of $g$,
we have
\begin{equation} \label{eqn:lem:progress:proximal_optimality}
	\ip{\nabla g(y)+\frac{1}{\mu}(y-p)}{u-y} \ge 0
	, \qquad
	g(u)-g(y) \ge \ip{\nabla g(y)}{u-y} \ge \frac{1}{\mu}\ip{p-y}{u-y}
. \end{equation}
Therefore,
\begin{align}
	\norm{p-u}^2
	&= \norm{y-u}^2 + \norm{y-p}^2 - 2\ip{p-y}{u-y}
	\notag\\
	&\ge \norm{y-u}^2 + \norm{y-p}^2 + 2\mu\left(g(y)-g(u)\right)
	\label{eqn:lem:progress:proximal_optimality_2}
. \end{align}
\par 
When $g(u)\le g(y)$,
the set $S = \X\cap\{z:g(z)\le g(y)\}$ contains $y$ and $u$.
We have the following lemma:
\begin{lemma}\label{lem:clipped-bound}
Let $p,y,u,v\in\R^d$,
$p\ne y$,
$y\ne v$,
$\mu,\lambda>0$,
$\kappa\ge1$,
and $0<\gamma<1$.
Suppose that
\begin{align*}
	0 \le g(u) &\le g(y),
	\qquad g(u) \ge \frac{\lambda}{2}\norm{u-v}^2,
	\\
	\norm{p-u}^2 &\ge \norm{y-u}^2 + \norm{y-p}^2 + 2\mu\left(g(y)-g(u)\right)
. \end{align*}
Let $S$ be a closed convex set containing $y$ and $u$,
and set
\[
	\eta = \min\left\{\gamma,\frac{\norm{y-p}}{2\kappa^{1/3}\norm{y-v}}\right\},
	\qquad x = \Pi_S\left\{(1-\eta)y+\eta v\right\}
. \]
Then
\begin{align}
	&~
	\norm{p-u} - \norm{x-u} + g(u)
	\notag\\
	\ge&~ \min\left\{
		\frac{\norm{y-p}}{2\kappa^{1/3}},
		\sqrt{\gamma(2-\gamma)\left(\norm{y-p}^2+2\mu(g(y)-g(u))\right)}
	\right\}
	+ g(u) - \gamma\sqrt{\frac{2g(u)}{\lambda}}
. \end{align}
\end{lemma}
And we have
\begin{align}
	&~
	\norm{p-u} - \norm{x-u} + g(u)
	\notag\\
	\ge&~ \min\left\{
		\frac{\norm{y-p}}{2\kappa^{1/3}},
		\sqrt{\gamma(2-\gamma)\left(\norm{y-p}^2+2\mu(g(y)-g(u))\right)}
	\right\}
	+ g(u) - \gamma\sqrt{\frac{2g(u)}{\lambda}}
	\label{eq:clipped-bound-application}
. \end{align}
\par 
If $g(u)\le g(y)/2$,
then $0<\gamma<1$ and $\mu = \norm{y-p}/G$ imply
\begin{align*}
	&\gamma(2-\gamma)\left(\norm{y-p}^2+2\mu(g(y)-g(u))\right)
	\\
	\ge&~ \gamma\left(\norm{y-p}^2+\mu g(y)\right)
	\\
	=&~ \frac{G}{4\kappa^{2/3}\left(G+1+\frac{3}{2\kappa^{1/3}}\right)}
	\left(1+\frac{1+\frac{3}{2\kappa^{1/3}}}{G}\right)\norm{y-p}^2
	\\
	=&~ \frac{\norm{y-p}^2}{4\kappa^{2/3}}
. \end{align*}
Substituting this into \eqref{eq:clipped-bound-application} gives
\begin{align*}
	&\norm{p-u} - \norm{x-u} + g(u)
	\\
	\ge&~ \frac{\norm{y-p}}{2\kappa^{1/3}} + g(u) - \gamma\sqrt{\frac{2g(u)}{\lambda}}
	\\
	=&~ \frac{\norm{y-p}}{2\kappa^{1/3}}
	+ \left(\sqrt{g(u)}-\frac{\gamma}{\sqrt{2\lambda}}\right)^2
	- \frac{\gamma^2}{2\lambda}
	\\
	\ge&~ \frac{\norm{y-p}}{2\kappa^{1/3}} - \frac{\norm{y-p}}{16\kappa^{1/3}}
	\\
	=&~ \frac{7}{16\kappa^{1/3}}\norm{y-p}
. \end{align*}
\par 
If $g(y)/2\le g(u)\le g(y)$,
Taking $u=p$ in the optimality condition \eqref{eqn:lem:progress:proximal_optimality} and using the Cauchy--Schwarz inequality yields
\begin{align*}
	G\norm{y-p}
	&= \frac{\norm{y-p}^2}{\mu}
	\\
	&\le \ip{\nabla g(y)}{p-y}
	\\
	&\le \norm{\nabla g(y)}\norm{y-p}
. \end{align*}
Since $y\ne p$,
we obtain $G\le\norm{\nabla g(y)}$.
Moreover,
global nonnegativity and smoothness imply the one-step improvement bound of the smoothed function:
\[
	0 \le g\left(y-\frac{\nabla g(y)}{L}\right)
	\le g(y) - \frac{\norm{\nabla g(y)}^2}{2L}
. \]
Consequently,
\begin{align*}
	\frac{\gamma^2}{\lambda}
	&= \frac{G^2}{16\lambda\kappa^{4/3}\left(G+1+\frac{3}{2\kappa^{1/3}}\right)^2}
	\\
	&\le \frac{Lg(y)}{8\lambda\kappa^{4/3}\left(G+1+\frac{3}{2\kappa^{1/3}}\right)^2}
	\\
	&\le \frac{\left(1+\frac{3}{2\kappa^{1/3}}\right)\norm{y-p}}{8\kappa^{1/3}\left(G+1+\frac{3}{2\kappa^{1/3}}\right)^2}
	\\
	&\le \frac{\norm{y-p}}{8(2\kappa^{1/3}+3)}
	\\
	&\le \frac{\norm{y-p}}{8\kappa^{1/3}}
. \end{align*}
Here we used $G^2\le\norm{\nabla g(y)}^2\le2Lg(y)$,
$L/\lambda\le\kappa$,
and $g(y) = (1+3/(2\kappa^{1/3}))\norm{y-p}$.
Taking this into \eqref{eq:clipped-bound-application}, and eliminating the $\min\{\dots\}$ term give
\begin{align*}
	&\norm{p-u} - \norm{x-u} + g(u)
	\\
	\ge&~ g(u) - \gamma\sqrt{\frac{2g(u)}{\lambda}}
	\\
	=&~ \frac{g(u)}{2}
	+ \left(\sqrt{\frac{g(u)}{2}}-\frac{\gamma}{\sqrt{\lambda}}\right)^2
	- \frac{\gamma^2}{\lambda}
	\\
	\ge&~ \frac{g(y)}{4} - \frac{\norm{y-p}}{8\kappa^{1/3}}
	\\
	=&~ \left(\frac14+\frac{1}{4\kappa^{1/3}}\right)\norm{y-p}
	\\
	\ge&~ \frac{\norm{y-p}}{2\kappa^{1/3}}
	\\
	\ge&~ \frac{7}{16\kappa^{1/3}}\norm{y-p}
. \end{align*}
The penultimate inequality uses $\kappa\ge1$.
\par 
If $g(u)\ge g(y)$,
then the triangle inequality and $\norm{x-y}\le\eta\norm{y-v}\le\norm{y-p}/(2\kappa^{1/3})$ give
\begin{align*}
	\norm{p-u} - \norm{x-u} + g(u)
	&\ge g(u) - \norm{x-p}
	\\
	&\ge g(y) - \norm{y-p} - \norm{x-y}
	\\
	&\ge \left(1+\frac{3}{2\kappa^{1/3}}-1-\frac{1}{2\kappa^{1/3}}\right)\norm{y-p}
	\\
	&= \frac{\norm{y-p}}{\kappa^{1/3}}
	\\
	&\ge \frac{7}{16\kappa^{1/3}}\norm{y-p}
. \end{align*}
The desired bound therefore holds in all three cases.
\end{proof}

\subsection{Proof of \texorpdfstring{\Cref{lem:clipped-bound}}{Lemma \getrefnumber{lem:clipped-bound}}}\label{app:clipped-bound}
\begin{proof}
Since $u\in S$,
nonexpansiveness of projection and the triangle inequality give
\begin{align*}
	\norm{x-u}
	&\le \norm{(1-\eta)y+\eta v-u}
	\\
	&= \norm{(1-\eta)(y-u)+\eta(v-u)}
	\\
	&\le (1-\eta)\norm{y-u} + \eta\norm{u-v}
. \end{align*}
Combining this with the assumed lower bound \eqref{eqn:lem:progress:proximal_optimality_2} yields
\begin{align*}
	&\norm{p-u} - \norm{x-u}
	\\
	\ge&~ \sqrt{\norm{y-u}^2+\norm{y-p}^2+2\mu(g(y)-g(u))}
	\\
	&\quad - (1-\eta)\norm{y-u} - \eta\norm{u-v}
	\\
	\ge&~ \min\left\{
		\frac{\norm{y-p}}{2\kappa^{1/3}},
		\sqrt{\gamma(2-\gamma)\left(\norm{y-p}^2+2\mu(g(y)-g(u))\right)}
	\right\} - \gamma\norm{u-v}
	\\
	\ge&~ \min\left\{
		\frac{\norm{y-p}}{2\kappa^{1/3}},
		\sqrt{\gamma(2-\gamma)\left(\norm{y-p}^2+2\mu(g(y)-g(u))\right)}
	\right\} - \gamma\sqrt{\frac{2g(u)}{\lambda}}
. \end{align*}
The second inequality follows from the following lemma:
\begin{lemma}\label{lem:geometry}
Let $y,u,v\in\R^d$,
$A,R>0$,
and $0<\gamma<1$.
Set
\[
	\eta = \min\{\gamma,R/\norm{y-v}\},
	\qquad\text{with }\eta = \gamma\text{ if }y=v
. \]
In particular,
\[
	\norm{u-v} \ge \left|\norm{y-v}-\norm{y-u}\right|,
	\qquad 0 < \eta \le \gamma,
	\qquad \eta\norm{y-v} = R\quad\text{if }\eta<\gamma
. \]
Then
\begin{align}
	&\sqrt{\norm{y-u}^2+A^2} - (1-\eta)\norm{y-u} + (\gamma-\eta)\norm{u-v}
	\notag\\
	\ge&~ \min\{R,A\sqrt{\gamma(2-\gamma)}\}
	\label{eq:geometry}
. \end{align}
\end{lemma}
And
\[
	A^2 = \norm{y-p}^2 + 2\mu\left(g(y)-g(u)\right) > 0,
	\qquad R = \frac{\norm{y-p}}{2\kappa^{1/3}}
. \]
The last inequality uses $g(u)\ge\lambda\norm{u-v}^2/2$.
Adding $g(u)$ proves the claim.
\end{proof}

\subsection{Proof of \texorpdfstring{\Cref{lem:geometry}}{Lemma \getrefnumber{lem:geometry}}}\label{app:geometry}
\begin{proof}
For $0\le a\le1$,
the Cauchy--Schwarz inequality gives
\begin{align*}
	a\norm{y-u} + A\sqrt{1-a^2}
	&\le \sqrt{\norm{y-u}^2+A^2}\sqrt{a^2+1-a^2}
	\\
	&= \sqrt{\norm{y-u}^2+A^2}
. \end{align*}
Hence
\begin{equation}\label{eq:geometry-cs}
	\sqrt{\norm{y-u}^2+A^2} - a\norm{y-u} \ge A\sqrt{1-a^2}
. \end{equation}
If $\eta=\gamma$,
taking $a=1-\gamma$ proves \eqref{eq:geometry}.
It remains to consider $\eta<\gamma$,
for which $\eta\norm{y-v}=R$.

First suppose that $R\le A\sqrt{\gamma(2-\gamma)}$.
If $\norm{y-u}\ge\norm{y-v}$,
then
\begin{align*}
	&\sqrt{\norm{y-u}^2+A^2} - (1-\eta)\norm{y-u} + (\gamma-\eta)\norm{u-v}
	\\
	\ge&~ \sqrt{\norm{y-u}^2+A^2} - (1-\eta)\norm{y-u}
	+ (\gamma-\eta)\left(\norm{y-u}-\norm{y-v}\right)
	\\
	=&~ \sqrt{\norm{y-u}^2+A^2} - \norm{y-u}
	+ \gamma\left(\norm{y-u}-\norm{y-v}\right) + R
	\\
	\ge&~ R
. \end{align*}
Here we used the reverse triangle inequality and $\eta\norm{y-v}=R$.
If $\norm{y-u}\le\norm{y-v}$ and $\eta\le\gamma/2$,
then
\begin{align*}
	&\sqrt{\norm{y-u}^2+A^2} - (1-\eta)\norm{y-u} + (\gamma-\eta)\norm{u-v}
	\\
	\ge&~ \sqrt{\norm{y-u}^2+A^2} - (1-\eta)\norm{y-u}
	+ (\gamma-\eta)\left(\norm{y-v}-\norm{y-u}\right)
	\\
	=&~ \sqrt{\norm{y-u}^2+A^2} - \norm{y-u}
	+ (\gamma-2\eta)\left(\norm{y-v}-\norm{y-u}\right) + R
	\\
	\ge&~ R
. \end{align*}
The first inequality again uses the reverse triangle inequality,
and the last uses $\gamma-2\eta\ge0$.
If $\norm{y-u}\le\norm{y-v}$ and $\gamma/2<\eta<\gamma$,
then
\begin{align*}
	&\eta^2(2-2\eta+\gamma) - \gamma(2-\gamma)(2\eta-\gamma)
	\\
	=&~ (\gamma-\eta)\left(2\eta^2+(2-\gamma)(\gamma-\eta)\right)
	\\
	\ge&~ 0
. \end{align*}
Since $0<2\eta-\gamma<1$ and $R\le A\sqrt{\gamma(2-\gamma)}$,
this implies
\begin{align*}
	A\sqrt{(2\eta-\gamma)(2-2\eta+\gamma)}
	&\ge \frac{A\sqrt{\gamma(2-\gamma)}(2\eta-\gamma)}{\eta}
	\\
	&\ge \frac{R(2\eta-\gamma)}{\eta}
. \end{align*}
Therefore,
\begin{align*}
	&\sqrt{\norm{y-u}^2+A^2} - (1-\eta)\norm{y-u} + (\gamma-\eta)\norm{u-v}
	\\
	\ge&~ \sqrt{\norm{y-u}^2+A^2} - (1-2\eta+\gamma)\norm{y-u}
	+ (\gamma-\eta)\norm{y-v}
	\\
	\ge&~ A\sqrt{(2\eta-\gamma)(2-2\eta+\gamma)} + (\gamma-\eta)\norm{y-v}
	\\
	\ge&~ \frac{R(2\eta-\gamma)}{\eta} + \frac{R(\gamma-\eta)}{\eta}
	\\
	=&~ R
. \end{align*}
The first inequality uses the reverse triangle inequality,
the second uses \eqref{eq:geometry-cs} with $a=1-2\eta+\gamma$,
and the third uses the preceding bound and $\eta\norm{y-v}=R$.

Finally,
suppose that $R\ge A\sqrt{\gamma(2-\gamma)}$ and $y\ne v$.
The preceding argument with radius $A\sqrt{\gamma(2-\gamma)}$ gives the lower bound $A\sqrt{\gamma(2-\gamma)}$ at
\[
	\eta = \min\left\{\gamma,\frac{A\sqrt{\gamma(2-\gamma)}}{\norm{y-v}}\right\}
. \]
The case $\eta=\gamma$ gives the same lower bound at the other endpoint.
Since the left side of \eqref{eq:geometry} is affine in $\eta$,
the bound holds whenever
\[
	\min\left\{\gamma,\frac{A\sqrt{\gamma(2-\gamma)}}{\norm{y-v}}\right\}
	\le \eta \le \gamma
. \]
The actual choice $\eta = \min\{\gamma,R/\norm{y-v}\}$ satisfies these inequalities,
which proves the claim.
\end{proof}

\subsection{Proof of \texorpdfstring{\Cref{lem:scalar}}{Lemma \getrefnumber{lem:scalar}}}\label{app:scalar}

\begin{proof}
Fix $q>0$. The function $\Lambda(q,\cdot)$ is positive and continuous on $[0,1)$ and diverges as $\delta\uparrow1$, and it therefore attains its minimum. Logarithmic differentiation yields
\[
\frac{\partial}{\partial\delta}\log\Lambda(q,\delta)=\frac{4q\delta}{1+q\delta^2}+\frac{2\delta}{1-\delta^2}-\frac{1}{q+\delta},
\]
and multiplication by the positive quantity $(1+q\delta^2)(1-\delta^2)(q+\delta)$ yields $P_q(\delta)$. We have $P_q(0)=-1$ and $P_q(1/\sqrt3)>0$. On $[0,1/\sqrt3]$, the derivative
\[
P_q'(\delta)=2(2-3\delta^2)q^2+(2+6\delta-4\delta^3)q+6\delta
\]
is positive. On $[1/\sqrt3,1)$, the first two terms of $P_q$ are positive and $3\delta^2-1\ge0$. Hence $P_q$ has a unique root $\ds(q)\in(0,1/\sqrt3)$, it is negative on $[0,\ds(q))$ and positive on $(\ds(q),1)$, and $\Lambda(q,\cdot)$ decreases on $[0,\ds(q)]$ and increases on $[\ds(q),1)$. The point $\ds(q)$ is consequently the unique minimizer.

For every $\delta\in[0,1)$,
\[
\frac{\partial}{\partial q}\log\Lambda(q,\delta)=-\frac{q(2-\delta^3)+\delta}{q(q+\delta)(1+q\delta^2)}<0.
\]
For $q_1<q_2$, this inequality yields $\phi(q_2)\le\Lambda(q_2,\ds(q_1))<\Lambda(q_1,\ds(q_1))=\phi(q_1)$, and $\phi$ is strictly decreasing. The function $\phi$ is the minimum of the jointly continuous function $\Lambda$ over $\delta$ in the compact interval $[0,1/\sqrt3]$, and it is therefore continuous. Finally, $\phi(q)\le\Lambda(q,0)=q^{-2}$, and the bounds $(1+q\delta^2)^2\ge1$, $1-\delta^2\le1$, and $q+\delta\le q+1$ yield $\phi(q)\ge1/(q(q+1))$.
\end{proof}

\subsection{Proof of \texorpdfstring{\Cref{cor:squared}}{Corollary \getrefnumber{cor:squared}}}\label{app:asymptotics}
\begin{proof}
We first prove the upper bound.
Since $\phi$ is decreasing and $\phi(\qs(\lambda))=\lambda$,
every pair $(q,\delta)$ satisfies
\[
	\phi(q)\le\Lambda(q,\delta)\le\lambda
	\implies
	\qs(\lambda)\le q
. \]
Taking $q=\lambda^{-1/2}$ and $\delta=0$ gives
\[
	\Lambda(\lambda^{-1/2},0)=\lambda
, \]
and hence $\qs(\lambda)\le\lambda^{-1/2}$.
Let $K=3\sqrt3/2$.
For $\lambda\ge K^2$, take $q=K/\lambda$ and $\delta=1/\sqrt3$.
Direct computation gives
\[
	\frac Kq-\Lambda\left(q,\frac1{\sqrt3}\right)
	=
	\frac{9\sqrt3-6-q}{6(q+1/\sqrt3)}
	\ge0
,
\]
where the inequality uses $q\le1/K<9\sqrt3-6$.
Thus $\Lambda(K/\lambda,1/\sqrt3)\le\lambda$, so $\qs(\lambda)\le K/\lambda$.
Therefore
\[
	\Cs(\lambda)=1+\qs(\lambda)
	\le
	1+\min\left\{\lambda^{-1/2},\frac K\lambda\right\}
. \]
The $\min$ branches when $\lambda\le K^2$.
\par 
We next establish the two limits.
As $\lambda\to0^+$, we have $\qs(\lambda)\to\infty$.
The bounds in \Cref{lem:scalar} give
\[
	\frac q{q+1}\le q^2\phi(q)\le1
. \]
Taking $q=\qs(\lambda)$ and using the equality $\phi(\qs(\lambda))=\lambda$,
we obtain
\[
	1 = \lim_{\qs(\lambda) \to \infty} \frac{\qs(\lambda)}{\qs(\lambda)+1}
	\le
	\lim_{\lambda \to 0^+}
	\qs(\lambda)^2 \lambda
	\le
	1
. \]
Therefore,
\[
	\lim_{\lambda \to 0^+}
	\sqrt\lambda\,\Cs(\lambda)
	=
	\lim_{\lambda \to 0^+}
	\sqrt\lambda+\sqrt{\lambda\qs(\lambda)^2}
	= 1
. \]
\par 
As $\lambda\to\infty$, we have $\qs(\lambda)\to0$.
We first show that $\lim_{q\to0}\ds(q)=1/\sqrt3$.
The polynomial $P_q$ converges uniformly on $[0,1]$
to $3\delta^2-1$.
Since $P_q(\ds(q))=0$, this convergence implies
\[
	\bigl|3\ds(q)^2-1\bigr|
	\le
	\sup_{\delta\in[0,1]}
	\bigl|P_q(\delta)-(3\delta^2-1)\bigr|
	=0
. \]
Together with $\ds(q)\in[0,1]$, this proves the claimed limit.
Consequently,
\[
	\lim_{q \to 0}
	q\phi(q)
	=
	\frac{(1+q\ds(q)^2)^2}
	{(1-\ds(q)^2)(q+\ds(q))}
	=
	\frac{3\sqrt3}{2}
. \]
Substituting $q=\qs(\lambda)$ gives
\[
	\lim_{\lambda \to \infty}
	\lambda\bigl(\Cs(\lambda)-1\bigr)
	=
	\lim_{\qs(\lambda) \to 0}
	\qs(\lambda)\phi(\qs(\lambda))
	=
	\frac{3\sqrt3}{2}
. \]
This completes the proof.
\end{proof}

\subsection{Proof of \texorpdfstring{\Cref{lem:lower}}{Theorem \getrefnumber{thm:lower}}}\label{app:lower}
\begin{proof}
Put $q=q_*(\lambda)$ and $\delta=\delta_*(q)$.
Let $C=1+q$ be the targeted competitive ratio, $a=h/(1-\delta),b=a^2/(q(1+\delta))$ be the parameters of the proposed algorithm,
and $h=1+q\delta^2$ be a frequent notation in the proof.
They depend only on $\lambda$.
By the previous section proving the upper bound,
\[
	\lambda = \frac{h^2}{q(1-\delta^2)(q+\delta)},
	P_q(\delta) = 0,
	0 < \delta < \frac{1}{\sqrt{3}}
, \]
where $\delta$ is the unique root of $P_q(\delta) = 2\delta(2 - \delta^2)q^2 + \delta(2 + 3\delta - \delta^3)q + 3\delta^2 - 1 = 0$ in $(0,1/\sqrt3)$.
Fix $0\le R<C$.
We will construct finite integers $L\ge1$ and $T_0\ge2$ at the end, using only $\lambda,R$.
For now,
take arbitrary integers $L\ge1$ and $T\ge2$,
set $d=T+L+1$,
and fix any online algorithm $\mathcal A$ in this dimension and horizon.
All requests and feasible decisions in the constructed instance belong to $\R^d$.
Let $\cost(\x_t)=\sum_{t=1}^Tf_t(\x_t)+1/2\|\x_t-\x_{t-1}\|^2$ denote the total cost of a path $\x_0,\x_1,\dots,\x_T$ under the constructed instance.
\par 
We first construct a reference path $\{X_t\}_{t=0}^T$ such that every online path $\{Y_t\}_{t=0}^T$ generated by $\mathcal A$ satisfies
\begin{equation}\label{eq:online-goal}
	\E[\ALG_T]
	= \E[\cost(Y_t)]
	\ge \cost(X_t)
. \end{equation}
The reference cost will be deterministic even though $X_t$ is random.
Let $\{\e_t\}_{t=0}^{T+L}$ be the orthonormal coordinate vectors of $\R^d$,
and $\{\sigma_t\}_{t=0}^{T+L}$ be independent uniform signs in $\{-1,1\}$.
Sample all these signs before play,
and define
\begin{equation} \label{eq:def:z_t}
	\z_t = \begin{cases}
		\e_0 ,& \text{ if } t=0 \\
		\mu \z_{t-1} + \sqrt{1-\mu^2}\sigma_t \e_t ,& \text{ otherwise}
	\end{cases}
. \end{equation}
Here we define
\begin{equation}\label{eq:lower-correlation}
	r = \frac{q\delta}{a}
	,\quad
	G = \frac{b}{\delta}
	,\quad
	\mu
	= 1-Gr
	= -\frac{C\delta^2}{1-\delta^2}
. \end{equation}


To verify $-1<\mu<0$,
evaluate $P_q$ at $1/\sqrt{q+2}$.
Grouping the even and odd powers gives
\begin{align*}
	P_q\left(\frac{1}{\sqrt{q+2}}\right)
	&= \frac{2q^2}{\sqrt{q+2}}\left(2-\frac{1}{q+2}\right)
		+ \frac{2q}{\sqrt{q+2}}
		+ \frac{3(q+1)}{q+2} - \frac{q}{(q+2)^2} - 1
	\\
	&= \frac{2q^2}{\sqrt{q+2}}\left(2-\frac{1}{q+2}\right)
		+ \frac{2q}{\sqrt{q+2}}
		+ \frac{3(q+1)(q+2)-q-(q+2)^2}{(q+2)^2}
	\\
	&= \frac{2q^2}{\sqrt{q+2}}\left(2-\frac{1}{q+2}\right)
		+ \frac{2q}{\sqrt{q+2}} + \frac{2(q+1)^2}{(q+2)^2}
	\\
	&> 0
. \end{align*}
Previous section has shown that $P_q$ is negative before its unique root and positive afterward on $(0,1)$.
Hence
\[
	\delta < \frac{1}{\sqrt{q+2}}
	\implies
	C\delta^2 < 1-\delta^2
. \]
Together with the expression for $\mu$ in \eqref{eq:lower-correlation}, this proves $\mu\in(-1,0)$.
Since $\z_{t-1}$ lies in the span of $\e_0,\ldots,\e_{t-1}$,
it is orthogonal to $\e_t$.
Induction gives
\[
	\begin{cases}
		\norm{\z_0} = 0 \\
		\norm{\z_t}^2 = \mu^2\norm{\z_{t-1}}^2 + 1-\mu^2
	\end{cases}
	\implies
	\norm{\z_t}^2 = 1, \text{ for all } t\ge0
. \]
Thus
\begin{equation} \label{eq:direction-inner-products}
	\ip{\z_s}{\z_t}
	= \mu\ip{\z_s}{\z_{t-1}}
	= \cdots = \mu^{t-s}
	, \text{ for all } 0\le s<t
. \end{equation}
Independence and the zero mean of $\sigma_{t+1}$ also give
\begin{equation}\label{eq:conditional-direction}
	\E\left[ \z_{t+1} \Given \sigma_1,\dots,\sigma_t \right] = \mu \z_t
. \end{equation}
Define the reference path by
\begin{equation} \label{eq:def:X_t}
	X_t = \begin{cases}
		0 ,& \text{ if } t=0 \\
		X_{t-1} - \frac{\z_t}{r} ,& \text{ otherwise}
	\end{cases}
. \end{equation}
For the nonterminal rounds,
set
\[
	f_t(\x)
	= \begin{cases}
		\max\left\{
			\frac{\lambda}{2}\norm{\x-\vb_t}^2,
			G\ip{\z_t}{\x-\vb_t}\right
		\} ,& \text{ if } t < T \\
		\frac{\lambda}{2}\norm{\x-X_T+\frac{\z_T}{\lambda r}}^2 ,& \text{ otherwise}
	\end{cases}
, \]
where $\vb_t=X_t-\z_t$.
In case of $t<T$, these functions are convex and nonnegative,
have unique minimizers $\vb_t$ with value zero,
and satisfy the required quadratic-growth inequality.
At $X_t$,
the two $\max$ branches in $t<T$ case have values $\lambda/2$ and $G$ respectively.
To identify the active branch,
compute
\begin{align*}
	\frac{G}{\lambda}
	&= \frac{a^2}{q\delta(1+\delta)}
	\frac{q(1-\delta^2)(q+\delta)}{h^2}
	= \frac{a^2(1-\delta)(q+\delta)}{\delta h^2}
	\notag\\
	&= \frac{q+\delta}{\delta(1-\delta)}
	= 1 + \frac{q+\delta^2}{\delta(1-\delta)}
	> 1
. \end{align*}
The third equality uses $a=h/(1-\delta)$.
By the above inequality,
the affine branch is strictly active there,
so
\begin{equation}\label{eq:reference-gradient}
	f_t(X_t) = G
	, \quad
	\nabla f_t(X_t) = G\z_t
	,
	\text{ for all } 1\le t<T
. \end{equation}
In the case of $t=T$,
$f_t(\x)$ is a nonnegative $\lambda$-QG cost with minimizer $X_T-\z_T/(\lambda r)$,
and
\begin{equation}\label{eq:terminal-gradient}
	f_T(X_T) = \frac{1}{2\lambda r^2}
	,\quad
	\nabla f_T(X_T) = \frac{\z_T}{r}
. \end{equation}
The input depends only on the signs sampled before play,
not on the learner's decisions.
\par 
We now prove \eqref{eq:online-goal}.
For any path $Y_0=0,Y_1,\dots,Y_T$,
set $\Delta_t=Y_t-X_t$,
so $\Delta_0=0$ and
\[
	Y_t-Y_{t-1} = -\frac{\z_t}{r} + \Delta_t-\Delta_{t-1}
. \]
The movement-cost difference is exactly
\[
	\frac12\norm{Y_t-Y_{t-1}}^2 - \frac12\norm{X_t-X_{t-1}}^2
	=
	-\frac1r\ip{\z_t}{\Delta_t-\Delta_{t-1}} + \frac12\norm{\Delta_t-\Delta_{t-1}}^2
. \]
Convexity of $f_t(\cdot)$, \eqref{eq:reference-gradient} and \eqref{eq:terminal-gradient} give
\begin{align*}
	f_t(Y_t)-f_t(X_t)
	&\ge
	\ip{\nabla f_t(X_t)}{Y_t-X_t}
	=
	G\ip{\z_t}{\Delta_t}
	, \text{ for all } t<T,
	\\
	f_T(Y_T)-f_T(X_T)
	&\ge
	\ip{\nabla f_T(X_T)}{Y_T-X_T}
	=
	\frac1r\ip{\z_T}{\Delta_T}
. \end{align*}
Therefore,
\begin{align}
	&~
	\cost(Y_t)-\cost(X_t)
	\notag\\
	\ge&~ G\sum_{t=1}^{T-1}\ip{\z_t}{\Delta_t}
	+ \frac1r\ip{\z_T}{\Delta_T}
	- \frac1r\sum_{t=1}^T\ip{\z_t}{\Delta_t-\Delta_{t-1}}
	+ \frac12\sum_{t=1}^T\norm{\Delta_t-\Delta_{t-1}}^2
	\notag\\
	=&~ G\sum_{t=1}^{T-1}\ip{\z_t}{\Delta_t}
	+ \frac1r\ip{\z_T}{\Delta_T}
	- \frac1r\sum_{t=1}^T\ip{\z_t}{\Delta_t}
	+ \frac1r\sum_{t=1}^T\ip{\z_t}{\Delta_{t-1}}
	+ \frac12\sum_{t=1}^T\norm{\Delta_t-\Delta_{t-1}}^2
	\notag\\
	=&~ \sum_{t=1}^{T-1}\ip{(G-\frac{1}{r})\z_t+\frac{1}{r}\z_{t+1}}{\Delta_t}
	+ \frac12\sum_{t=1}^T\norm{\Delta_t-\Delta_{t-1}}^2
	\label{eq:online-variation}
. \end{align}
The last step shifts the index in the last inner-product sum,
uses $\Delta_0=0$,
and cancels the two terms involving $\Delta_T$.

Let $\F_t$ be the sigma-field generated by $\sigma_1,\dots,\sigma_t$ and $\xi$.
For $t<T$,
the requests revealed through round $t$ depend only on $\sigma_1,\dots,\sigma_t$.
Hence $\Delta_t$ is $\F_t$-measurable for an online path.
Using \eqref{eq:conditional-direction} and $\mu=1-Gr$,
we have
\begin{align*}
	\E\left[ \ip{(G-\frac{1}{r})\z_t+\frac{1}{r}\z_{t+1}}{\Delta_t} \right]
	&=
	\E\left[ \ip{\E[(G-\frac{1}{r})\z_t+\frac{1}{r}\z_{t+1} \given \F_t] }{\Delta_t} \right]
	\\
	&= \E\left[ \ip{(G-\frac1r+\frac{\mu}{r})\z_t}{\Delta_t} \right]
	\\
	&= \E\left[ \ip{(G-\frac1r+\frac{1-Gr}{r})\z_t}{\Delta_t} \right]
	\\
	&= 0
. \end{align*}
Therefore,
\[
	\E[ \cost(Y_t) ] - \E[ \cost(X_t) ]
	= \E\left[ \frac12\sum_{t=1}^T\norm{\Delta_t-\Delta_{t-1}}^2 \right]
	\ge 0
. \]
Taking expectations in \eqref{eq:online-variation} yields
\begin{align}
	\E_{\sigma,\xi}[ \ALG_T ]
	&= \E_{\sigma,\xi}\left[ \cost(Y_t) \right]
	\notag\\
	&\ge \E_{\sigma,\xi}[\cost(X_t)]
	\notag\\
	&= (T-1)\left(G+\frac{1}{2r^2}\right)
	+ \frac{1}{2\lambda r^2} + \frac{1}{2r^2}
	\notag\\
	&= (T-1)\left(G+\frac{1}{2r^2}\right) + \frac{1+1/\lambda}{2r^2}
	\label{eq:online-lower-bound}
. \end{align}
Here the second equality is by \eqref{eq:reference-gradient} and \eqref{eq:terminal-gradient}.
This argument still applies if the learner knows $T$ and the input distribution.
\par 
We next construct a feasible offline path $\{U_t\}_{t=0}^T$ with small total cost.
Since the offline optimum minimizes over all feasible paths,
this will give $\OPT_T\le\cost(U_t)$ for each fixed input.
Introduce the weight
\[
	\beta = \frac{q(1-\delta^2)}{C}
. \]
It is positive,
and
\[
	1-\beta
	= \frac{C-q(1-\delta^2)}{C}
	= \frac{1+q-q+q\delta^2}{C}
	= \frac{h}{C}
	> 0
. \]
Hence $0<\beta<1$.
Use the next $L$ directions to define
\begin{equation} \label{eq:finite-comparator}
	U_t = \begin{cases}
		0 ,& \text{ if } t=0 \\
		X_t-E_t^{(L)} ,& \text{ for all } 1\le t<T \\
		X_T ,& \text{ if } t=T
	\end{cases}
, \end{equation}
where
\[
	E_t^{(L)} = \frac1r\sum_{j=1}^{L}\beta^j \z_{t+j}
. \]
Each $U_t$ belongs to the same $\R^d$ as the requests,
and $U_0=\x_0$.
Therefore,
\[
	\operatorname{OPT}_T
	\le \cost(U_t)
. \]

To compute the cost of $\{U_t\}_{t=0}^T$,
we first evaluate its untruncated counterpart
\[
	\Uh_t = X_t-E_t
	,\text{ where }
	E_t = \frac1r\sum_{j=1}^{\infty}\beta^j \z_{t+j},
. \]
To show the validation of this infinite series $\{\z_t\}_{t=0}^\infty$,
we embed $\R^d$ into the Hilbert space $\R^\infty$ satisfying square-summable sequences $\|\x\|^2=\sum_{t=0}^\infty x_t^2<\infty$,
and extend coordinate basis to $\{\e_t\}_{t=0}^\infty$ and additional independent uniform signs $\{\sigma_t\}_{t=0}^\infty$.
Use the same recursion \eqref{eq:def:z_t} and \eqref{eq:def:X_t} to extend $\z_t$ and $X_t$,
and $f_t(\x)=\max\{ \lambda/2\norm{\x-\vb_t}^2, G\ip{\z_t}{\x-\vb_t}\}$ to extend $f_t(\cdot)$.
This extension does not change the finite-dimensional instance or the path $\{U\}_{t=0}^T$.
Since $0<\beta<1$ and $\norm{\z_t}=1$,
this series converges absolutely.
\par 
We consider evaluate the cost of $\{\Uh_t\}_{t=0}^T$.
To compute $f_t(\Uh_t) = \max\{G\ip{\z_t}{\Uh_t-\vb_t},\lambda/2\|\Uh_t-\vb_t\|^2\}$, we first compute the inner product $\ip{\z_t}{\Uh_t-\vb_t}$.
Since $\vb_t=X_t-\z_t,\Uh_t-\vb_t=\z_t-E_t$,
The affine branch equals
\[
	G\ip{\z_t}{\Uh_t-\vb_t}
	= G\ip{\z_t}{\z_t-E_t}
	= G(1-\ip{\z_t}{E_t})
. \]
For the inner product of $\z_t$ and $E_t$,
\eqref{eq:direction-inner-products} gives
\begin{align*}
	\ip{\z_t}{E_t}
	&= \frac1r\sum_{j=1}^{\infty}\beta^j\mu^j
	= \frac{\beta\mu}{r(1-\beta\mu)}
	\notag\\
	&= -\frac{q\delta^2}{rh}
	= -\frac{q\delta^2a}{q\delta h}
	= -\frac{a\delta}{h}
	\notag\\
	&= -\frac{\delta}{1-\delta}
. \end{align*}
Here the third equality is by
\[
	\beta\mu
	= \frac{q(1-\delta^2)}{C}\left(-\frac{C\delta^2}{1-\delta^2}\right)
	= -q\delta^2
	,\quad
	1-\beta\mu = h
. \]
Therefore,
\[
	G\ip{\z_t}{\Uh_t-\vb_t}
	=
	\frac{G}{1-\delta}
. \]
Then we compute $\lambda/2\|\Uh_t-\vb_t\|^2$.
\[
	\|\Uh_t-\vb_t\|^2
	=
	\norm{\z_t-E_t}^2
	= 1 - 2\ip{\z_t}{E_t} + \norm{E_t}^2
	= 1 + \frac{2\delta}{1-\delta} + \norm{E_t}^2
. \]
For the squared norm term $\|E_t\|^2$,
absolute convergence permits grouping the diagonal terms and the two equal off-diagonal sums:
\begin{align*}
	\norm{E_t}^2
	&= \frac{1}{r^2}\sum_{i=1}^{\infty}\sum_{j=1}^{\infty}\beta^{i+j}\mu^{|i-j|}
	\\
	&= \frac{1}{r^2}\left(\sum_{i=1}^{\infty}\beta^{2i}
	+ 2\sum_{i=1}^{\infty}\sum_{\ell=1}^{\infty}\beta^{2i+\ell}\mu^{\ell}\right)
	\\
	&= \frac{\beta^2}{r^2(1-\beta^2)}\left(1+2\sum_{\ell=1}^{\infty}(\beta\mu)^{\ell}\right)
	\\
	&= \frac{\beta^2}{r^2(1-\beta^2)}\left(1+\frac{2\beta\mu}{1-\beta\mu}\right)
	\\
	&= \frac{\beta^2(1+\beta\mu)}{r^2(1-\beta^2)(1-\beta\mu)}
	= \left(\frac{\beta}{r}\right)^2 \frac{1}{1-\beta^2} \frac{1+\beta\mu}{1-\beta\mu}
. \end{align*}
We next substitute the parameters $\beta=(q(1-\delta^2))/C,r=(q\delta)/a,a=h/(1-\delta)$ into the above equation.
The two factors needed for the substitution are
\begin{align*}
	\frac{\beta}{r}
	&= \frac{q(1-\delta^2)}{C}\frac{a}{q\delta}
	= \frac{a(1-\delta^2)}{C\delta}
	= \frac{h(1+\delta)}{C\delta},
	\\
	1-\beta^2
	&= \frac{C^2-q^2(1-\delta^2)^2}{C^2}
	= \frac{(C-q(1-\delta^2))(C+q(1-\delta^2))}{C^2}
	= \frac{h(1+2q-q\delta^2)}{C^2}
. \end{align*}
Together with $\beta\mu=-q\delta^2$ and $1-\beta\mu=h$,
these identities give
\begin{equation} \label{eq:gap-norm-rational}
	\norm{E_t}^2
	= \frac{h^2(1+\delta)^2}{C^2\delta^2}
	\frac{(1-q\delta^2)C^2}{h^2(1+2q-q\delta^2)}
	= \frac{(1+\delta)^2(1-q\delta^2)}{\delta^2(1+2q-q\delta^2)}
. \end{equation}
In particular,
$\norm{E_t}^2$ is deterministic and independent of $t$.
To further evaluate $((1+\delta)^2(1-q\delta^2))/(\delta^2(1+2q-q\delta^2))$,
we use the equation $P_q(\delta) = 2\delta(2 - \delta^2)q^2 + \delta(2 + 3\delta - \delta^3)q + 3\delta^2 - 1 = 0$ and expand
\[
	(1+\delta)^2(1-q\delta^2) \cdot (1-\delta)^2
	=
	1-2\delta^2+\delta^4 - q(\delta^2-2\delta^4+\delta^6),
, \]
and
\[
	\delta(1+2q-q\delta^2) \cdot (2q+\delta+\delta^3)
	=
	(4\delta-2\delta^3)q^2 + (2\delta+2\delta^2+\delta^4-\delta^6)q + \delta^2+\delta^4
. \]
Subtracting these expansions yields
\begin{align*}
	&~
	(1+\delta)^2(1-q\delta^2) \cdot (1-\delta)^2
	- \delta(1+2q-q\delta^2) \cdot (2q+\delta+\delta^3)
	\\
	=&~ 1-3\delta^2 - (2\delta+3\delta^2-\delta^4)q - (4\delta-2\delta^3)q^2
	\\
	=&~ -P_q(\delta)= 0. \end{align*}
Dividing by the positive denominator $\delta^2(1-\delta)^2(1+2q-q\delta^2)$ gives
\[
	\frac{(1+\delta)^2(1-q\delta^2)}{\delta^2(1+2q-q\delta^2)}
	= \frac{2q+\delta+\delta^3}{\delta(1-\delta)^2}
. \]
Therefore, $\|E_t\|^2=(2q+\delta+\delta^3)/(\delta(1-\delta)^2)$, and
\begin{align*}
	\|\Uh_t-\vb_t\|^2
	&= 1 - 2\ip{\z_t}{E_t} + \norm{E_t}^2
	\\
	&= 1 + \frac{2\delta}{1-\delta}
	+ \frac{2q+\delta+\delta^3}{\delta(1-\delta)^2}
	\\
	&= \frac{\delta(1-\delta)^2+2\delta^2(1-\delta)+2q+\delta+\delta^3}{\delta(1-\delta)^2}
	\\
	&= \frac{2(q+\delta)}{\delta(1-\delta)^2}
. \end{align*}
By $\lambda=\Lambda_0(q,\delta)=h^2/(q(1-\delta^2)(q+\delta))$
the quadratic branch at $\Uh_t$ equals
\begin{align*}
	\frac{\lambda}{2}\norm{\Uh_t-\vb_t}^2
	&= \lambda \cdot \frac{(q+\delta)}{\delta(1-\delta)^2}
	= \frac{h^2}{q(1-\delta^2)(q+\delta)} \cdot \frac{q+\delta}{\delta(1-\delta)^2}
	\\
	&= \frac{a^2}{q\delta(1-\delta^2)}
	= \frac{b}{\delta(1-\delta)}
	\\
	&= \frac{G}{1-\delta}
. \end{align*}
Combining the two $\max$ branches gives
\[
	f_t(\Uh_t) = \frac{G}{1-\delta}, \text{ for all } 1\le t<T
. \]
We next evaluate the moving cost $\|\Uh_t-\Uh_{t-1}\|^2$.
\begin{align*}
	\Uh_t-\Uh_{t-1}
	&= -\frac{\z_t}{r}-E_t+E_{t-1}
	\\
	&= -\frac{\z_t}{r} - \frac{1}{r} \sum_{j=1}^{\infty}\beta^j \z_{t+j} + \frac{1}{r} \sum_{j=1}^{\infty}\beta^j \z_{t-1+j}
	\\
	&= - \frac{1}{r\beta} \sum_{j=1}^{\infty}\beta^j \z_{t-1+j} + \frac{1}{r} \sum_{j=1}^{\infty}\beta^j \z_{t-1+j}
	\\
	&= (1-\frac{1}{\beta}) E_{t-1}
. \end{align*}
Since $\norm{E_{t-1}}^2=\norm{E_t}^2$,
its nonterminal one-round cost is
\begin{equation}\label{eq:offline-cost-before-simplification}
	f_t(\Uh_t) + \frac12\norm{\Uh_t-\Uh_{t-1}}^2
	= \frac{G}{1-\delta}
	+ \frac{(1-\beta)^2}{2\beta^2}\norm{E_t}^2
. \end{equation}
We simplify the two terms using the expressions for $\|E_t\|^2$ derived above.
$\|E_t\|^2=(2q+\delta+\delta^3)/(\delta(1-\delta)^2)$ and \eqref{eq:gap-norm-rational} implies
\begin{align*}
	\frac1{\delta^2}
	&= q+\frac{1+2q-q\delta^2}{(1+\delta)^2}\norm{E_t}^2
	\\
	&= \frac{1+q}{(1+\delta)^2}\norm{E_t}^2
	+q\left(1+\frac{1-\delta}{1+\delta}\norm{E_t}^2\right)
	\\
	&= \frac{1+q}{(1+\delta)^2}\norm{E_t}^2
	+q\left(1+\frac{2q+\delta+\delta^3}{\delta(1-\delta^2)}\right)
	\\
	&= \frac{1+q}{(1+\delta)^2}\norm{E_t}^2
	+\frac{2q(q+\delta)}{\delta(1-\delta^2)}
.
\end{align*}
Thus,
\[
	\frac{\norm{E_t}^2}{(1+\delta)^2}
	= \frac1{1+q}
	\left(
	\frac1{\delta^2}
	-\frac{2q(q+\delta)}{\delta(1-\delta^2)}
	\right)
. \]
Since
\[
	\frac{\lambda}{2}\norm{\z_t-E_t}^2
	= \frac{G}{1-\delta},
	\qquad
	\norm{\z_t-E_t}^2
	= 1+\frac{2\delta}{1-\delta}+\norm{E_t}^2,
\]
by $\|E_t\|^2=(2q+\delta+\delta^3)/(\delta(1-\delta)^2)$ we obtain
\begin{align}
	G
	&= \frac{\lambda}{2}
	\left(1+\delta+(1-\delta)\norm{E_t}^2\right)
	= \frac{\lambda}{2}
	\left(
	1+\delta+\frac{2q+\delta+\delta^3}{\delta(1-\delta)}
	\right)
	\notag\\
	&= \frac{\lambda(q+\delta)}{\delta(1-\delta)}
	= \frac{b}{\delta}
	= \frac{a^2}{q\delta(1+\delta)}
	\label{eq:G-simplified}
,
\end{align}
where we used $(q+\delta)\lambda=(1-\delta)b$ and $b=a^2/[q(1+\delta)]$.
Substituting this expression and $(1-\beta)/\beta=a/[q(1+\delta)]$ into \eqref{eq:offline-cost-before-simplification} gives
\begin{align*}
	f_t(\Uh_t) + \frac12\norm{\Uh_t-\Uh_{t-1}}^2
	&= \frac{G}{1-\delta}
	+\frac{a^2}{2q^2}\frac{\norm{E_t}^2}{(1+\delta)^2}
	\\
	&= \frac{G}{1-\delta}
	+\frac{a^2}{2q^2(1+q)}
	\left(
	\frac1{\delta^2}
	-\frac{2q(q+\delta)}{\delta(1-\delta^2)}
	\right)
	\\
	&= \frac{G}{1-\delta}
	+\frac{a^2}{2q^2(1+q)\delta^2}
	-\frac{q+\delta}{(1+q)(1-\delta)}
	\frac{a^2}{q\delta(1+\delta)}
	\\
	&= \frac{G}{1-\delta}
	+\frac{a^2}{2q^2(1+q)\delta^2}
	-\frac{(q+\delta)G}{(1+q)(1-\delta)}
	\\
	&= \frac{(1+q)-(q+\delta)}{(1+q)(1-\delta)}G
	+\frac{a^2}{2q^2(1+q)\delta^2}
	\\
	&= \frac1{1+q}
	\left(G+\frac{a^2}{2q^2\delta^2}\right),
\end{align*}
where the fourth equality uses \eqref{eq:G-simplified}.
Finally, recalling $1/r=a/(q\delta)$ and $C=1+q$, we conclude that
\begin{equation}\label{eq:auxiliary-cost}
	f_t(\Uh_t) + \frac12\norm{\Uh_t-\Uh_{t-1}}^2
	= \frac{G+1/(2r^2)}{C}
	, \text{ for all } 1\le t<T
. \end{equation}
\par 
We now bound the truncation error and the cost of the two corrected endpoints in \eqref{eq:finite-comparator}.
At a nonterminal round, $U_t-\vb_t=\z_t-E_t^{(L)}$ and $\Uh_t-\vb_t=\z_t-E_t$,
the finite and infinite sums satisfy
\begin{align*}
	\norm{E_t},\norm{E_t^{(L)}}
	&\le \frac1r\sum_{j=1}^{\infty}\beta^j
	= \frac{\beta}{r(1-\beta)},
	\\
	\|U_t-\Uh_t\|
	\le
	\norm{E_t-E_t^{(L)}}
	&\le \frac1r\sum_{j=L+1}^{\infty}\beta^j
	= \frac{\beta^{L+1}}{r(1-\beta)}
. \end{align*}
At a nonterminal round, $U_t-\vb_t=\z_t-E_t^{(L)}$ and $\Uh_t-\vb_t=\z_t-E_t$,
so both points lie within distance $1+\beta/(r(1-\beta))$ of $\vb_t$.
Define
\[
	M_f = \max\left\{\lambda\left(1+\frac{\beta}{r(1-\beta)}\right),G\right\}
	. \]
The difference between the quadratic-branch values is at most
\begin{align*}
	\frac{\lambda}{2}\left|\norm{U_t-\vb_t}^2-\norm{\Uh_t-\vb_t}^2\right|
	&\le \frac{\lambda}{2}
	\left(\norm{U_t-\vb_t}+\norm{\Uh_t-\vb_t}\right)\norm{U_t-\Uh_t}
	\\
	&\le \frac{\lambda}{2}
	\left(2 \|\z_t\| + \|E_t\| + \|E^{(L)}_t\|\right)\norm{U_t-\Uh_t}
	\\
	&\le \lambda\left(1+\frac{\beta}{r(1-\beta)}\right)
	\frac{\beta^{L+1}}{r(1-\beta)}
. \end{align*}
The difference between the affine-branch values is at most
\[
	G\left|\ip{\z_t}{U_t-\Uh_t}\right|
	\le G \norm{\z_t} \norm{U_t-\Uh_t}
	\le \frac{G\beta^{L+1}}{r(1-\beta)}
. \]
The absolute difference of two maxima is bounded by the maximum of the absolute differences of their corresponding branches.
Consequently,
\begin{equation}\label{eq:hitting-truncation}
	|f_t(U_t)-f_t(\Uh_t)|
	\le
	\frac{M_f\beta^{L+1}}{r(1-\beta)}
. \end{equation}
For $2\le t<T$,
the two movement vectors satisfy
\begin{align*}
	\norm{U_t-U_{t-1}}
	&= \norm{-\z_t/r-E_t^{(L)}+E_{t-1}^{(L)}}
	\\
	&\le \frac1r+\frac{2\beta}{r(1-\beta)}
	= \frac{1+\beta}{r(1-\beta)},
	\\
	\norm{\Uh_t-\Uh_{t-1}}
	&= \norm{-\z_t/r-E_t+E_{t-1}}
	\\
	&\le \frac1r+\frac{2\beta}{r(1-\beta)}
	= \frac{1+\beta}{r(1-\beta)},
	\\
	\norm{(U_t-U_{t-1})-(\Uh_t-\Uh_{t-1})}
	&\le \norm{E_t-E_t^{(L)}} + \norm{E_{t-1}-E_{t-1}^{(L)}}
	\\
	&\le \frac{2\beta^{L+1}}{r(1-\beta)}
. \end{align*}
Applying the difference-of-squares identity and the Cauchy--Schwarz inequality to these movement vectors gives
\begin{align*}
	&~
	\frac12\left|\norm{U_t-U_{t-1}}^2-\norm{\Uh_t-\Uh_{t-1}}^2\right|
	\\
	&= \left(\norm{U_t-U_{t-1}}+\norm{\Uh_t-\Uh_{t-1}}\right) \frac12\left|\norm{U_t-U_{t-1}}-\norm{\Uh_t-\Uh_{t-1}}\right|
	\\
	&\le \frac12\left(\frac{2(1+\beta)}{r(1-\beta)}\right) \left(\frac{2\beta^{L+1}}{r(1-\beta)}\right)
	\\
	&= \frac{2(1+\beta)\beta^{L+1}}{r^2(1-\beta)^2}
. \end{align*}
Together with \eqref{eq:auxiliary-cost} and \eqref{eq:hitting-truncation},
this proves
\begin{align*}
	&~
	f_t(U_t)+\frac12\norm{U_t-U_{t-1}}^2
	\\
	\le&~
	f_t(\Uh_t)+\frac12\norm{\Uh_t-\Uh_{t-1}}^2
		+ |f_t(U_t)-f_t(\Uh_t)| + \frac12\left|\norm{U_t-U_{t-1}}^2-\norm{\Uh_t-\Uh_{t-1}}^2\right|
	\\
	\le&~
	\frac{G+1/(2r^2)}{C}
	+ \frac{\beta^{L+1}}{r(1-\beta)}\left(M_f+\frac{2(1+\beta)}{r(1-\beta)}\right)
, \end{align*}
for all $2 \le t < T$.
In the first round,
\[
	\norm{U_1-\vb_1} \le 1+\frac{\beta}{r(1-\beta)},
	\quad
	\norm{U_1-U_0} \le \frac1r+\frac{\beta}{r(1-\beta)}
	= \frac1{r(1-\beta)}
. \]
Define the resulting hitting-cost bound by
\[
	H_0 = \max\left\{
		\frac{\lambda}{2}\left(1+\frac{\beta}{r(1-\beta)}\right)^2,
		G\left(1+\frac{\beta}{r(1-\beta)}\right)
	\right\}
. \]
Then
\[
	f_1(U_1)+\frac12\norm{U_1-U_0}^2
	\le H_0+\frac1{2r^2(1-\beta)^2}
. \]
In the last round,
$U_T=X_T$ and $U_T-U_{T-1}=-\z_T/r+E_{T-1}^{(L)}$,
so
\[
	f_T(U_T)+\frac12\norm{U_T-U_{T-1}}^2
	\le \frac1{2\lambda r^2}+\frac1{2r^2(1-\beta)^2}
. \]
Summing the interior and endpoint bounds yields
\begin{align*}
	\operatorname{OPT}_T
	&\le \cost(U)
	= \sum_{t=1}^T f_t(U_t)+\frac12\norm{U_t-U_{t-1}}^2
	\\
	&\le (T-2)\left(
	\frac{G+1/(2r^2)}{C}
	+ \frac{\beta^{L+1}}{r(1-\beta)}\left(M_f+\frac{2(1+\beta)}{r(1-\beta)}\right)
	\right)
	\\
	&\quad + H_0+\frac1{2\lambda r^2}+\frac1{r^2(1-\beta)^2}
. \end{align*}
Here $M_f,H_0$ depend only on $\lambda$,
and both the requests and the comparator are restricted to the space of original dimension.
\par 
The reference and auxiliary costs are now explicit.
Choose a finite $L\ge1$ such that
\[
	R\frac{\beta^{L+1}}{r(1-\beta)} \left(M_f+\frac{2(1+\beta)}{r(1-\beta)}\right)
	<
	\frac{1}{4} (1 - \frac{R}{C}) (G+\frac{1}{2r^2})
. \]
Such an $L$ exists because $0<\beta<1$.
Next choose a finite $T_0\ge2$ such that
\[
	\frac{A+R\left(H_0+1/(2\lambda r^2)+1/(r^2(1-\beta)^2)\right)}{T_0}
	<
	\frac{1}{4} (1 - \frac{R}{C}) (G+\frac{1}{2r^2})
. \]
Both choices depend only on $\lambda,R$,
and are independent of the algorithm.

For every $T\ge T_0$,
the online and offline bounds imply
\begin{align*}
	&~
	\E_{\sigma,\xi}(\ALG_T)-R\E_{\sigma}(\operatorname{OPT}_T)
	\\
	\ge&~ (T-1) (G+\frac{1}{2r^2})
	\\&
	-RT\left(
		\frac{G+\frac{1}{2r^2}}{C}+\frac{\beta^{L+1}}{r(1-\beta)}
		\left(M_f+\frac{2(1+\beta)}{r(1-\beta)}\right)
	\right)
	- R\left(H_0+\frac1{2\lambda r^2}+\frac1{r^2(1-\beta)^2}\right)
	\\
	=&~ T (1 - \frac{R}{C}) (G+\frac{1}{2r^2})
	\\&
		-RT\frac{\beta^{L+1}}{r(1-\beta)}
		\left(M_f+\frac{2(1+\beta)}{r(1-\beta)}\right)
	- (G+\frac{1}{2r^2}) - R\left(H_0+\frac1{2\lambda r^2}+\frac1{r^2(1-\beta)^2}\right)
	\\
	>&~ \frac{1}{2} T (1 - \frac{R}{C}) (G+\frac{1}{2r^2})
	> 0
. \end{align*}
The first inequality drops the positive terminal term in \eqref{eq:online-lower-bound},
and replaces $T-2$ by $T$ in the nonnegative offline bound.
The second inequality follows from the choices of $L$ and $T_0$ above.

Therefore, for every $T\ge T_0$,
\[
	\frac{\E_{\xi}[\ALG_T]}{\OPT_T}
	\ge
	C = 1 + q_*(\lambda)
. \]
This means
\[
	\inf_{A\in\A} \sup_{\substack{d\in\D\\(f_1,\dots,f_T)\in\F}} \frac{\E[\ALG_T]}{\OPT_T}
	\ge
	1+q_*(\lambda)
. \]
\end{proof}

\end{document}